\documentclass{article}

\usepackage{iclr2027_conference,times}

\usepackage[T1]{fontenc}
\usepackage[utf8]{inputenc}
\usepackage{microtype}
\usepackage{graphicx}
\usepackage{booktabs}
\usepackage{amsmath}
\usepackage{multirow}
\usepackage{caption}
\usepackage{float}
\usepackage{placeins}
\usepackage{needspace}
\usepackage{wrapfig}
\usepackage{xcolor}
\usepackage{enumitem}
\usepackage[most]{tcolorbox}
\usepackage{url}
\usepackage{hyperref}

\title{Not All Flips Are Conformity:\\Decomposing Stance Convergence in Multi-Agent LLM Debate}

\newcommand{\bit}{$^{1}$}
\newcommand{\osaka}{$^{2}$}
\newcommand{\szu}{$^{3}$}

\author{\bit Xiqi Hao, \osaka Zengqing Wu\thanks{Corresponding authors. Our source code is available at \url{https://github.com/47hao030808/Not-All-Flips-Are-Conformity/}.}, \bit Yu-Xuan Qiu, \osaka Chuan Xiao, \bit Ruiqi Xu, \osaka Shuyuan Zheng, \szu Jianbin Qin\footnotemark[1]\\
  \bit Beijing Institute of Technology, Zhuhai, \osaka University of Osaka, \szu Shenzhen University\\
  \texttt{zengqing.wu@ist.osaka-u.ac.jp, qinjianbin@szu.edu.cn}
}

\iclrfinalcopy

\begin{document}

\maketitle
\lhead{Preprint}

\begin{abstract}
Multi-agent debate (MAD) among large language models (LLMs) is often judged by answer changes and convergence, yet these outcomes alone cannot establish that peers improved the answer. We study this gap with matched self-reflection, stance-only, and full-reasoning conditions. With five GPT-4o agents on MMLU-Pro disagreement questions, 37\% of agent-question pairs change even without peer information. Among pairs stable under this control, 29.3\% adopt a peer answer after seeing peer stances. These adoptions include corrections, new errors, and changes between wrong answers. The outcomes vary with correct-answer availability. When a correct peer answer is available, correction exceeds harmful adoption after accounting for peer support in the primary setting. Full reasoning improves individual accuracy but provides no additional majority-vote gain over self-reflection while using 57\% more tokens. Responses to constructed peer materials vary across GPT-4o, DeepSeek-V4-Flash, and Qwen3-Plus. An answer-agnostic intervention suppresses both harmful and beneficial peer adoption without a detectable accuracy gain. Together, the results show that convergence is an ambiguous signal of value and that debate must be judged by what peer information contributes to the final group answer.
\end{abstract}

\section{Introduction}

Multi-agent debate (MAD) among large language models (LLMs) is often assessed through answer changes and convergence. These observations do not establish that the peers helped. An agent may revise its answer when asked to reconsider without seeing any peer response. A change toward a peer may correct an error, introduce an error, or leave the answer wrong. Treating these outcomes as one signal obscures both the source of a revision and its value to the final group answer. More agreement need not mean a better decision.

We study these distinctions primarily with five GPT-4o agents that first answer each multiple-choice question without peer information. Each agent-question pair on a disagreement question is then evaluated under three matched conditions: self-reflection without peers, exposure to peer stances, and exposure to peer stances with reasoning. A change without peers is reconsideration instability. Peer adoption after stability under self-reflection is strict conformity, while adoption after stability under both self-reflection and stance-only exposure is reasoning-only peer adoption. These labels describe observable response patterns.

We ask three questions. What drives stance convergence beyond reconsideration without peers (RQ1)? How do peer materials change wrong-answer adoption (RQ2)? Can harmful peer adoption be predicted and mitigated (RQ3)? These questions connect the source of individual revisions to the value of the final group answer.

Our central finding is that both the baseline for answer change and the direction of peer adoption matter. In the primary MMLU-Pro setting, 36.98\% of 5,200 agent-question pairs from disagreement questions change under self-reflection alone. Among pairs stable in that control, 29.30\% adopt a peer answer under stance-only exposure. These peer adoptions include corrections, new errors, and switches between wrong answers. When a correct peer is available, beneficial adoption exceeds harmful adoption after standardizing for support of the target's initial answer. The value of convergence therefore depends on which alternatives peers make available and whether adopting them improves the eventual group decision.

Individual improvement also need not translate into a better group answer. On the disagreement questions, individual accuracy rises from 42.17\% under self-reflection to 45.87\% under full reasoning. Across all 4,900 questions, however, both conditions yield 77.35\% majority-vote accuracy, while full reasoning uses 57\% more tokens. This result is specific to the homogeneous, single-round setting studied here. It shows why the value of debate must be assessed at the level of the decision it is meant to improve, as well as at the level of individual revisions.

Our main contribution is a matched account of stance convergence that connects the source and correctness of answer changes to the group decision. We evaluate it with GPT-4o across four multiple-choice benchmarks and with selected checks on DeepSeek-V4-Flash and Qwen3-Plus. These model families come from different providers and release generations. Their different stability profiles show the value of a common measurement framework across models, with rates that remain model specific. Our controlled peer-material analysis examines how complete peer presentations change wrong-answer adoption. The intervention analysis then tests whether identifying likely adopters is sufficient to improve accuracy.

\section{Related Work}
\label{sec:related}

\paragraph{Multi-agent debate and self-reflection.}
Multi-agent debate lets LLM instances challenge and refine one another, supporting reasoning and evaluation~\citep{du2023improving,liang-etal-2024-encouraging,zhang-etal-2024-exploring,chen-etal-2024-reconcile,estornell2024multilllm,chan2023chateval}. Yet debate does not consistently outperform single-agent baselines~\citep{smit2024should,wang-etal-2024-rethinking-bounds,wynn2025talk,zhang2025stop,yang2025revisiting}. \citet{choi2025debate} find that initial majority voting explains much of the gain attributed to multi-round debate. They also study belief updating and interventions that give correct responses greater influence. This work motivates measuring how individual revisions contribute to the final group answer. Our matched one-round design tracks individual C$\to$W, W$\to$C, and W$\to$W transitions under no-peer reconsideration, peer stances, and additional peer reasoning, and relates them to group majority accuracy across peer-support compositions. This joint analysis captures how corrections and errors can offset after aggregation. While iterative refinement can improve outputs, reconsideration without external feedback can also degrade them~\citep{madaan2023selfrefine,huang2024selfcorrect}. This finding motivates our self-reflection control. We also measure how correct-peer availability changes the opportunity for correction.

\paragraph{Conformity and sycophancy in LLMs.}
Peer-exposure studies, drawing parallels to Asch's conformity experiments~\citep{asch1956conformity}, find that peer stances increase answer changes and examine mitigations such as anonymization, agent pairing, and dynamic prompting~\citep{zhu-etal-2025-conformity,weng2025doaswedo,lin2025socialinfluence,choi-etal-2025-empirical,cho2025herd,choihk2025identity,yao2025peacemaker,pitre-etal-2025-consensagent}. Work on single-agent sycophancy and signal competition treats these as related phenomena~\citep{sharma2024sycophancy,wei2024synthetic,zhang2025signalcompetition}. \citet{baltaji2024persona} use private pre/post responses and discussion transcripts to distinguish conformity from opinion and persona inconstancy. Building on this distinction, we add a matched no-peer re-elicitation for each target. From a common Round~0 answer, we then measure adoption after exposure to peer stances and peer reasoning, paralleling the normative/informational distinction~\citep{deutsch1955normative}.

\paragraph{Information quality and persuasion.}
Semantic content can affect model reasoning even when logical structure is relatively stable~\citep{dasgupta2022content}, while adversarial text can mislead models by mimicking valid reasoning despite containing logical errors~\citep{turpin2023unfaithful,cui2025madspear}. Our information-gradient experiment compares GPT-4o's responses to stance-only, assertion-only, generic non-solving text, and plausible question-specific wrong reasoning under controlled peer exposure. The resulting contrasts measure responses to complete presentations that differ in content, length, and other surface properties.

\section{Methodology}
\label{sec:methodology}

\subsection{Counterfactual Decomposition}

We study a MAD setting with $N = 5$ language model agents independently answering each multiple-choice question in Round~0 without agent identity or peer information, producing initial answers $a_i^{(0)}$. Here, independently means without inter-agent communication, not statistical independence. In Round~1, each target on a disagreement question answers again under three parallel conditions starting from the same Round~0 answer. Peer adoption denotes a change to an answer provided by a peer. We decompose the observed response patterns using a matched no-peer condition and two peer-exposure conditions. Let $a_i^{(\text{sr})}$, $a_i^{(\text{so})}$, and $a_i^{(\text{r})}$ denote target $i$'s Round~1 answers under self-reflection, stance-only, and full reasoning conditions:

\begin{table}[H]
\centering
\caption{Operational event categories in the primary MMLU-Pro sample. Conditions are evaluated in parallel from the same Round~0 state.}
\label{tab:event_definitions}
\small
\renewcommand{\arraystretch}{1.12}
\setlength{\tabcolsep}{4pt}
\begin{tabular}{@{}p{0.25\textwidth}p{0.62\textwidth}r@{}}
\toprule
\textbf{Event category} & \textbf{Required response pattern} & $\boldsymbol{n}$ \\
\midrule
Reconsideration instability & $a_i^{(\mathrm{sr})}\neq a_i^{(0)}$. The target changes without peer information. & 1,923 \\
Strict conformity & $a_i^{(\mathrm{sr})}=a_i^{(0)}$, $a_i^{(\mathrm{so})}\neq a_i^{(0)}$, and $a_i^{(\mathrm{so})}$ matches a peer's Round~0 answer. & 960 \\
Reasoning-only peer adoption & $a_i^{(\mathrm{sr})}=a_i^{(0)}$, $a_i^{(\mathrm{so})}=a_i^{(0)}$, $a_i^{(\mathrm{r})}\neq a_i^{(0)}$, and $a_i^{(\mathrm{r})}$ matches a peer's Round~0 answer. & 487 \\
\bottomrule
\end{tabular}
\end{table}

\begin{figure*}[t]
\centering
\includegraphics[width=\linewidth,pagebox=cropbox,clip]{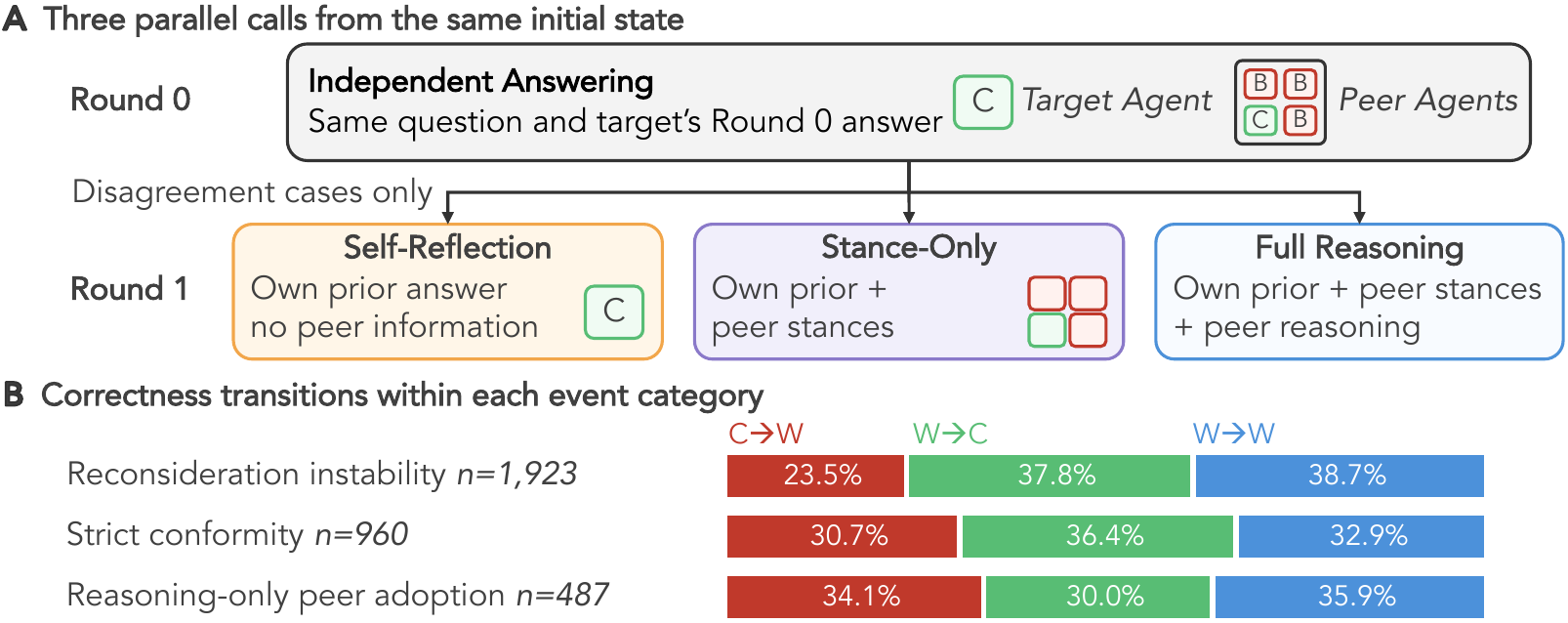}
\caption{Matched parallel conditions and correctness transitions within the operational event categories (GPT-4o, MMLU-Pro). Each bar sums to 100\% within its named event group. C and W denote correct and wrong answers.}
\label{fig:decomposition}
\end{figure*}

Table~\ref{tab:event_definitions} defines the three event categories. We assign self-reflection changes first, leaving $5{,}200-1{,}923=3{,}277$ self-reflection-stable pairs eligible for strict conformity. Reasoning-only peer adoption additionally requires stability under stance-only exposure. Other response patterns, including stance-only changes to answers absent from peer stances after stable self-reflection, form the residual ``Other'' category ($n=143$). Pairs unchanged in all three conditions are stable ($n=1{,}687$). The five categories partition all 5,200 pairs. Separately, 295 stance-only and 38 full-reasoning changes end at answers absent from the peers' Round~0 stances. These marginal counts can overlap the partition categories. The labels describe observable response patterns. Figure~\ref{fig:decomposition} shows the parallel design and correctness transitions within the three event categories.

Some peer-exposure studies operationalize conformity through post-exposure alignment \cite{zhu-etal-2025-conformity,cho2025herd}, while private pre/post designs already document both conformity and inconstancy \cite{baltaji2024persona}. Our operational definition addresses a narrower attribution question by requiring stability under a matched self-reflection control before labeling a peer-associated change as conformity. The strict-conformity rate uses self-reflection-stable samples as the denominator:
\begin{equation}
    \text{Strict-Conformity Rate}
    =
    \frac{N_{\mathrm{strict}}}{N_{\text{SR-stable}}}.
\end{equation}
Here, $N_{\mathrm{strict}}$ counts strict-conformity pairs and $N_{\text{SR-stable}}$ counts self-reflection-stable pairs (960 and 3,277, respectively, in the primary sample).

Each peer-adoption event is classified by its correctness transition. C$\to$W changes a correct Round~0 answer to a wrong answer, W$\to$C changes a wrong answer to a correct answer, and W$\to$W ends wrong after starting wrong. We refer to C$\to$W as harmful peer adoption and W$\to$C as beneficial peer adoption. W$\to$W changes the selected answer without changing correctness. Self-reflection names the no-peer condition, and its flip rate measures reconsideration instability. The deployment analysis measures peer adoption without conditioning on a self-reflection response (Section~\ref{sec:rq3}).

\begin{figure}[!t]
\centering
\includegraphics[width=\linewidth,pagebox=cropbox,clip]{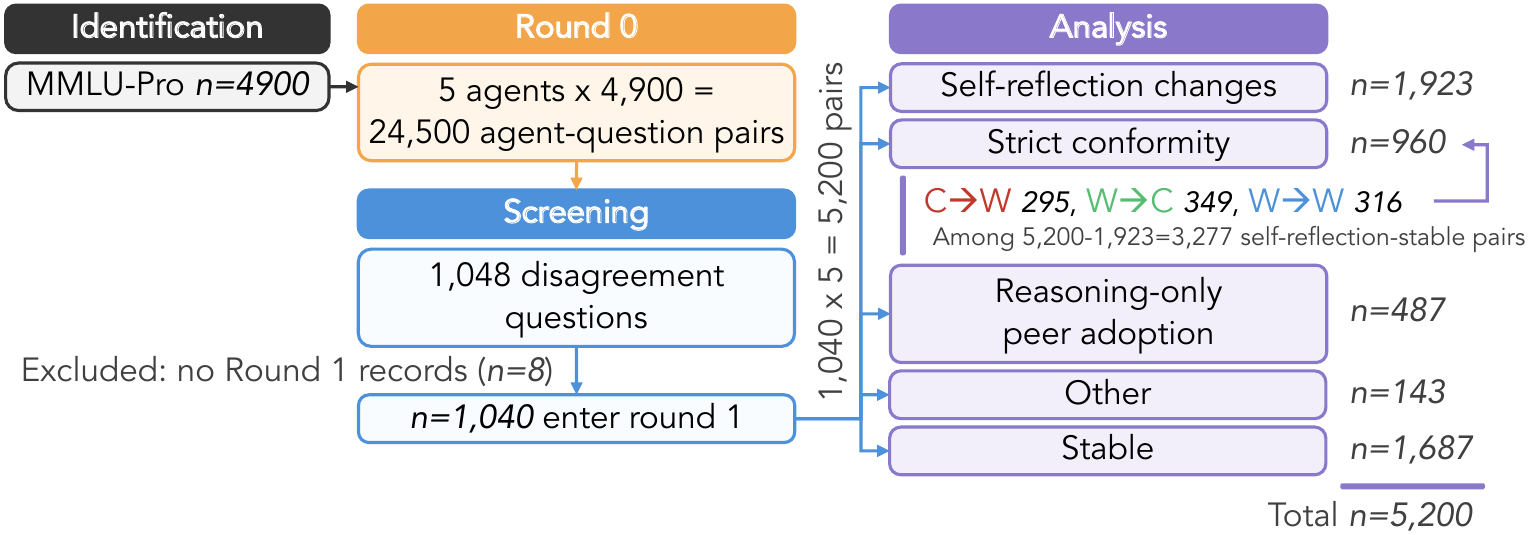}
\caption{Sample flow and counterfactual partition for MMLU-Pro. From 4,900 Round~0 questions, 1,048 exhibit inter-agent disagreement. Eight have no corresponding Round~1 records, leaving 1,040 questions and 5,200 agent--question pairs (five agents per question). Responses under the three parallel conditions are classified into five mutually exclusive categories totaling 5,200. The strict-conformity rate uses the 3,277 self-reflection-stable pairs as its denominator. Reasoning-only peer adoption additionally requires stability under stance-only exposure. C and W denote correct and wrong answers.}
\label{fig:sample_flow}
\end{figure}

\subsection{Experimental Setup}

In Round~0, each of the 5 agents independently answers the question under an identical, information-isolated prompt. The system records each agent's answer, reasoning trace, and implicit confidence (derived from output token logprobs, Appendix~\ref{app:implicit_confidence}). In Round~1, each agent-question pair on a disagreement question is evaluated in separate calls under all three conditions (self-reflection, stance-only, full reasoning), enabling within-pair counterfactual comparisons. The full prompt templates appear in Appendix~\ref{app:prompt_templates}.

\paragraph{Datasets.}
We use MMLU-Pro~\citep{wang2024mmlupro} as the primary benchmark and GPQA-Diamond~\citep{rein2023gpqa} for cross-dataset validation. Additional replications on BBEH and AGIEval are reported in Appendix~\ref{app:cross_dataset}.

\paragraph{Models.}
All 5 agents use GPT-4o (temperature 0). Cross-model replications with DeepSeek-V4-Flash and Qwen3-Plus (2025-12-01) are in Appendix~\ref{app:cross_model}. Question-specific wrong-reasoning materials for the information-gradient experiment (Section~\ref{sec:rq2}) are generated by GPT-5.4, with a Qwen3-Max generator variant in Appendix~\ref{app:rq2_145}.

\paragraph{Scale.}
From MMLU-Pro, we sample 4,900 questions, yielding 24,500 agent--question pairs in Round~0. Of the 1,048 questions with Round~0 disagreement, eight lack Round~1 data. The remaining 1,040 questions yield 5,200 agent--question pairs. Figure~\ref{fig:sample_flow} traces this sample construction and the resulting partition. From GPQA-Diamond, 105 disagreement questions yield 525 agent-question pairs.

\section{RQ1: What Drives Stance Convergence?}
\label{sec:rq1}

Among the answer changes observed during MAD, how much is attributable to peer-associated change beyond reconsideration instability under information isolation?

\subsection{Answer Flip Rates and Strict Conformity}

Table~\ref{tab:flip_rates} reports raw answer flip rates for each condition. The self-reflection condition already produces flip rates of 36.98\% on MMLU-Pro and 39.24\% on GPQA-Diamond, indicating that more than one-third of observations would change even without peer information. The stance-only condition adds about 7--10 percentage points over self-reflection, and full reasoning adds about a further 4 points. This ordering holds in these GPT-4o runs. Cross-model departures from it, including Qwen3-Plus, are discussed in Appendix~\ref{app:cross_model}.

\begin{table}[H]
\centering
\begin{minipage}[t]{0.49\textwidth}
\vspace{0pt}
\centering
\caption{Answer flip rates across conditions.}
\label{tab:flip_rates}
\setlength{\tabcolsep}{2pt}
\begin{tabular}{@{}lcc@{}}
\toprule
\textbf{Condition} & \textbf{MMLU-Pro} & \textbf{GPQA} \\
\midrule
Self-reflection & 36.98\% & 39.24\% \\
Stance-only & 46.85\% & 46.48\% \\
Full reasoning & 50.56\% & 51.05\% \\
\bottomrule
\end{tabular}
\end{minipage}
\hfill
\begin{minipage}[t]{0.49\textwidth}
\vspace{0pt}
\raggedright
\captionsetup{justification=raggedright,singlelinecheck=false}
\caption{Support-standardized harmful (H) and \mbox{beneficial} (B) peer adoption (\%).}
\label{tab:support_standardized}
\setlength{\tabcolsep}{3pt}
\begin{tabular}{@{}lrrr@{}}
\toprule
\textbf{Exposure} & \textbf{Std. H} & \textbf{Std. B} & \textbf{H$-$B (pp)} \\
\midrule
Stance-only & 24.36 & 33.12 & $-$8.75 \\
Full reasoning & 32.09 & 40.42 & $-$8.33 \\
\bottomrule
\end{tabular}
\end{minipage}
\end{table}

Applying the strict conformity definition (self-reflection stable, stance-only changed to a peer's answer) yields a strict-conformity rate of 29.30\% (95\% question-cluster bootstrap CI [27.5, 31.2]) on MMLU-Pro ($n=960/3{,}277$, 1,040 question clusters) and 29.15\% (95\% question-cluster bootstrap CI [24.2, 34.4]) on GPQA ($n=93/319$, 105 question clusters). These intervals use 20,000 question-cluster bootstrap resamples (seed 20260921). This is lower than the stance-only flip rate (46.85\%). Of 2,436 stance-only flips, 1,342 (55.1\%) also change under self-reflection, though not necessarily to the same answer. Among the same 2,436 flips, 960 (39.4\%) satisfy the strict conformity definition.

The self-reflection flip rate itself depends on sampling variability and on the re-elicitation instruction. On all 198 GPQA-Diamond questions, three identical information-isolated Round~0 calls disagree on 22.4\% of within-question response pairs. Re-eliciting from a fixed prior answer changes 8.8\% of answers under a neutral answer-again instruction and 16.7\% under an explicit reconsideration instruction, a paired difference of 7.91 pp (question-clustered 95\% CI [4.38, 11.62], Appendix~\ref{app:gpqa_direction}).

\Needspace{0.32\textheight}
\subsection{Directional Analysis}

\par\medskip
\noindent
Table~\ref{tab:decomposition} reports C$\to$W, W$\to$C, and W$\to$W within each event source. In strict conformity, 295 events are C$\to$W, 349 are W$\to$C, and 316 are W$\to$W. Among self-reflection changes, the corresponding counts are 452, 727, and 744.

\begin{table}[H]
\centering
\begin{minipage}[t]{0.49\textwidth}
\vspace{0pt}
\raggedright
\captionsetup{justification=raggedright,singlelinecheck=false}
\caption{Decomposition and directional analysis\protect\\(\%, MMLU-Pro, GPT-4o).}
\label{tab:decomposition}
\setlength{\tabcolsep}{1pt}
\begin{tabular}{@{}lrrrr@{}}
\toprule
\textbf{Source} & $n$ & \textbf{C$\to$W} & \textbf{W$\to$C} & \textbf{W$\to$W} \\
\midrule
\shortstack[l]{Self-reflection\\changes} & 1,923 & 23.50 & 37.81 & 38.69 \\
\shortstack[l]{Strict\\conformity} & 960 & 30.73 & 36.35 & 32.92 \\
\shortstack[l]{Reasoning-only\\peer adoption} & 487 & 34.09 & 29.98 & 35.93 \\
\bottomrule
\end{tabular}
\end{minipage}
\hfill
\begin{minipage}[t]{0.49\textwidth}
\vspace{0pt}
\centering
\caption{Oracle-controlled correct-signal support gradient ($n=250$ per condition).}
\label{tab:correct_signal_gradient}
\setlength{\tabcolsep}{2pt}
\begin{tabular}{@{}lrr@{}}
\toprule
\shortstack[l]{\textbf{Peer}\\\textbf{composition}} & \shortstack{\textbf{Accuracy}\\\textbf{(\%)}} & \shortstack{\textbf{$\Delta$ prev.}\\\textbf{(pp)}} \\
\midrule
No correct signal (0/4) & 5.6 & -- \\
Correct minority (1/4) & 14.0 & $+$8.4 \\
Balanced (2/4) & 61.2 & $+$47.2 \\
Correct majority (3/4) & 89.2 & $+$28.0 \\
\bottomrule
\end{tabular}
\end{minipage}
\end{table}

The direction of peer influence depends on correction opportunity and support. Of 1,811 initially wrong, self-reflection-stable targets, 891 (49.2\% [46.9, 51.5]) have no correct peer, and 205 (11.3\% [9.9, 12.9]) face a correct-peer majority. We compare harmful peer adoption (H) among initially correct targets with beneficial peer adoption (B) among initially wrong targets that have at least one correct peer. These populations differ in how many peers support the target's initial answer. To compare them at a common support distribution, we standardize both rates over support counts of 0/1/2/3 using the pooled frequencies of these strata in the two opportunity populations. As Table~\ref{tab:support_standardized} shows, standardized beneficial peer adoption exceeds harmful peer adoption under both exposure conditions. H$-$B is $-$8.75 percentage points [95\% CI: $-$13.44, $-$4.27] under stance-only exposure and $-$8.33 [$-$13.17, $-$3.60] under full reasoning. These intervals are question-clustered. Full support-stratum rates appear in Appendix~\ref{app:support_standardization}.

An oracle-controlled experiment uses ground-truth labels to construct peer answers with known correctness. We use this support-gradient experiment to test how correct-peer availability changes accuracy. Accuracy rises from 5.6\% with no correct peer to 14.0\% with one, 61.2\% with two, and 89.2\% with three ($n=250$ per condition, Table~\ref{tab:correct_signal_gradient}). Relative to the preceding condition, the increases are $+$8.4 percentage points [95\% CI: $+$5.2, $+$12.0], $+$47.2 [$+$41.2, $+$53.2], and $+$28.0 [$+$22.4, $+$33.6], respectively. Intervals use paired question-cluster bootstrap resampling. The first contrast introduces a correct peer while preserving the 3:1 wrong majority. Later contrasts also change peer support. Appendix~\ref{app:correct_signal_gradient} details this ground-truth-based construction.

\subsection{Robustness}

Appendices~\ref{app:gpqa_direction}--\ref{app:cross_dataset} report directional outcomes for eight model--benchmark combinations, covering GPT-4o, DeepSeek-V4-Flash, and Qwen3-Plus on MMLU-Pro and GPQA-Diamond, plus GPT-4o on BBEH and AGIEval. W$\to$C counts are at least as large as C$\to$W counts in all eight combinations, but only the primary GPT-4o MMLU-Pro cluster-bootstrap interval for the W$\to$C share excludes 50\%. The other intervals are wide. Self-reflection flip rates also vary substantially, from 11.91\% for DeepSeek-V4-Flash to 74.07\% for Qwen3-Plus on MMLU-Pro. Both reconsideration changes and strict-conformity events occur in every evaluated model and benchmark combination, although their prevalence varies markedly. The evidence for a directional imbalance is strongest in the primary sample. On GPQA-Diamond, peer order has little effect, whereas prompt wording, agent count, and temperature alter the measured rates (Appendix~\ref{app:sensitivity}). MMLU-Pro subject-level rates appear in Appendix~\ref{app:per_category}.

\subsection{Net Accuracy Effect of MAD}

Across 5,200 agent--question pairs from the 1,040 disagreement questions, individual accuracy rises from 36.88\% at Round~0 to 42.17\% under self-reflection, 44.48\% under stance-only exposure, and 45.87\% under full reasoning. Across all 4,900 questions, majority-vote accuracy is 76.22\% at Round~0, 77.18\% after stance-only exposure, and 77.35\% after either self-reflection or full reasoning. In paired question transitions, Round~0 to self-reflection produces 130 newly correct and 75 newly wrong majority answers. Self-reflection to full reasoning produces 107 of each. Thus, the full-reasoning condition yields no net majority-vote accuracy gain over self-reflection in this setting. The paired majority transitions measure the group outcome directly, while the directional event counts describe individual revisions.

Table~\ref{tab:accuracy_cost} places these accuracies against token cost over all 4,900 questions. Full-reasoning debate uses 57\% more tokens than self-reflection while both attain 77.35\% majority-vote accuracy, and their paired accuracy difference is 0.00 pp [95\% CI $-$0.59, $+$0.59]. Aggregation-only or self-reflection therefore gives a better observed accuracy--cost trade-off than applying full-reasoning exchange by default. Token accounting details are in Appendix~\ref{app:accuracy_cost}.

\begin{table}[H]
\centering
\caption{Accuracy--cost trade-off on all 4,900 MMLU-Pro questions. Token counts include five Round~0 calls for multi-instance conditions and the applicable Round~1 calls on retained disagreement questions.}
\label{tab:accuracy_cost}
\setlength{\tabcolsep}{3pt}
\begin{tabular}{lrrrr}
\toprule
\textbf{Condition} & \textbf{Accuracy (\%)} & \textbf{Tokens/question} & \textbf{$\times$ single} & \textbf{$\Delta$Acc. vs. aggregation} \\
\midrule
Single instance & 75.13 & 700 & 1.00 & $-$1.10 pp \\
Aggregation-only & 76.22 & 3,500 & 5.00 & -- \\
Self-reflection & 77.35 & 4,469 & 6.38 & $+$1.12 pp \\
Peer-stance-only debate & 77.18 & 4,522 & 6.46 & $+$0.96 pp \\
Full-reasoning debate & 77.35 & 7,019 & 10.03 & $+$1.12 pp \\
\bottomrule
\end{tabular}
\end{table}

The no-peer control and directional accounting show why convergence alone cannot measure peer benefit. Full reasoning changes additional individual answers but does not improve the majority vote over self-reflection in this setting.

\section{RQ2: How Do Peer Materials Change Wrong-Answer Adoption?}
\label{sec:rq2}

The absence of a net group gain in RQ1 leaves open how peer reasoning redirects individual answers. We examine the direction of reasoning-only peer adoption, then vary the presentation of a fixed wrong answer to study when peer explanations induce its adoption.

\subsection{Direction of Reasoning-Only Peer Adoption}

In MMLU-Pro, 487 agent--question pairs resist self-reflection and stance-only exposure but adopt a peer answer under full reasoning. Among them, 146 (29.98\%) change from wrong to correct, 166 (34.09\%) change from correct to wrong, and 175 (35.93\%) remain wrong. The split shows that reasoning-only peer adoption includes corrections, newly incorrect answers, and changes between wrong answers.

\subsection{The Four-Condition Information Gradient Experiment}

The five-agent debate reveals where peer reasoning accompanies answer changes, but each message comes with a particular mix of peer answers and support. To examine how the presentation of a wrong answer relates to adoption, a one-peer experiment holds the question and constructed wrong answer fixed. We use a susceptibility-selected sample of 500 GPT-4o target-agent/question anchors from 100 question clusters, selected from 126 questions with candidate full-reasoning correct-to-wrong events. This sampling focuses the presentation contrasts on candidate harmful-adoption cases. Its rates describe the selected anchors and do not estimate adoption in ordinary debate.

All four conditions use one peer, an ``answer again'' instruction, the same questions and options, a constructed wrong answer, and fixed GPT-generated peer materials. The conditions present this answer as a stance, an assertion of correctness, generic non-solving text, or a plausible question-specific wrong rationale. Representative peer materials appear in Appendix~\ref{app:reasoning_examples}. This controlled experiment isolates responses to constructed peer materials on selected anchors. Each condition also provides an assigned correct prior, which matches the target's actual Round~0 answer in 253 of the 500 anchors. The contrasts therefore describe responses to the assigned prior and fixed materials.

Table~\ref{tab:rq2_cross_target_full} (GPT-4o column) shows the pattern. Adoption is 64.4\% under stance-only, 64.8\% under assertion-only, 78.2\% under invalid/non-diagnostic reasoning, and 87.0\% under question-specific wrong reasoning. Adoption changes by $+0.4$ pp from stance-only to assertion-only [95\% CI $-5.8$, $+6.4$], then by $+13.4$ pp from assertion-only to invalid/non-diagnostic reasoning [$+7.8$, $+19.6$], and by $+8.8$ pp from invalid/non-diagnostic to question-specific wrong reasoning [$+2.2$, $+15.6$]. Question-specific wrong reasoning exceeds stance-only by $+22.6$ pp [$+15.6$, $+30.4$].

These incremental contrasts measure responses to the complete peer presentations. The materials are controlled but not length-matched: assertion-to-invalid adds reasoning-like structure and approximately 130 tokens, while invalid-to-question-specific-wrong adds approximately 100 tokens and possible presentation differences.

\begin{table}[H]
\centering
\caption{Wrong-answer adoption rates (\%, $n=500$ per condition and target). DeepSeek-V4 denotes DeepSeek-V4-Flash. All targets receive the same GPT-generated anchors and materials.}
\label{tab:rq2_cross_target_full}
\setlength{\tabcolsep}{5pt}
\begin{tabular}{lrrr}
\toprule
\textbf{Condition} & \textbf{GPT-4o} & \textbf{DeepSeek-V4} & \textbf{Qwen3-Plus} \\
\midrule
Stance-only & 64.4 & 20.0 & 23.4 \\
Assertion-only & 64.8 & 23.8 & 37.8 \\
Invalid/non-diagnostic & 78.2 & 11.8 & 36.4 \\
Question-specific wrong & 87.0 & 37.2 & 54.4 \\
\bottomrule
\end{tabular}
\end{table}

\begin{table}[H]
\centering
\caption{Paired adoption-rate contrasts (pp, $n=500$, 100 question clusters per target) with question-clustered 95\% CIs.}
\label{tab:rq2_target_contrasts}
\setlength{\tabcolsep}{5pt}
\begin{tabular}{lcc}
\toprule
\textbf{Target} & \textbf{Invalid$-$assertion} & \textbf{Wrong$-$invalid} \\
\midrule
GPT-4o & $+$13.4 [$+$7.8, $+$19.6] & $+$8.8 [$+$2.2, $+$15.6] \\
DeepSeek-V4-Flash & $-$12.0 [$-$17.8, $-$6.6] & $+$25.4 [$+$17.8, $+$33.0] \\
Qwen3-Plus & $-$1.4 [$-$5.6, $+$3.2] & $+$18.0 [$+$11.2, $+$25.2] \\
\bottomrule
\end{tabular}
\end{table}

Appendix~\ref{app:rq2_145} examines outcome-based selection and independent calls on 145 indices, while Appendix~\ref{app:rq2_original_full} examines another 500-anchor material protocol. The generic-text contrast varies across these analyses, while question-specific wrong rationales are associated with higher adoption in each.

Table~\ref{tab:rq2_cross_target_full} also extends the same fixed anchors and materials to DeepSeek-V4-Flash and Qwen3-Plus as answering models. Final accuracy per condition is reported in Appendix~\ref{app:cross_target_extensions}, Table~\ref{tab:rq2_cross_target_accuracy}. Table~\ref{tab:rq2_target_contrasts} reports the paired adoption contrasts with question-clustered intervals. The invalid-minus-assertion contrast is positive for GPT-4o ($+$13.4 pp), reverses direction for DeepSeek-V4-Flash ($-$12.0 pp), and is near zero for Qwen3-Plus ($-$1.4 pp), whereas wrong-reasoning-minus-invalid is positive for all three targets. The response-model comparison holds the GPT-4o anchors and materials fixed while changing the answering model. Between-target interaction contrasts are reported in Appendix~\ref{app:cross_target_extensions}. A complementary paired target-resistance experiment applies identical anonymous 3:1 wrong-majority pressure to GPT-4o and Qwen3-Plus on 14,055 common-valid cases (Appendix~\ref{app:target_resistance}). Qwen3-Plus is 2.14 pp less likely to abandon its correct answer. This result characterizes target-dependent response under a constructed fixed context.

Across these selected anchors, question-specific wrong rationales elicit more adoption than generic non-solving text for all three answering models, whereas the effect of generic text varies by target and protocol. The result identifies susceptibility to plausible wrong peer explanations, while the value of any given adoption still depends on the target's initial answer and the alternatives available to it.

\section{RQ3: Can Conformity Risk Be Predicted and Mitigated?}
\label{sec:rq3}

Because peer adoption can correct as well as corrupt an answer, intervention requires more than predicting who will change. We first compare prediction of adoption and harmful adoption (Section~\ref{sec:rq3_prediction}). We then test an intervention with oracle eligibility, which requires ground-truth correctness and self-reflection stability (Section~\ref{sec:rq3_diagnostic}). Finally, we evaluate intervention using answer-agnostic Round~0 features across all 5,200 pairs (Section~\ref{sec:rq3_deployment}). The diagnostic and deployment tasks have different labels and populations, so their AUCs are not directly comparable.

\subsection{Predicting Peer Adoption and Oracle-Defined Risk}
\label{sec:rq3_prediction}

Round~0 confidence differs only modestly by correctness and behavioral outcome. The full group summaries are in Appendix~\ref{app:rq3_feature_ablation}. The answer-agnostic model predicts stance-only peer adoption across all 5,200 agent--question pairs (AUC = 0.804), but does not distinguish whether adoption is harmful or beneficial.

To evaluate harmful peer adoption directly, we refit the same nested question-grouped logistic model using the nine deployable Round~0 features listed in Appendix~\ref{app:deployable_features} and a stance-only C$\to$W peer-adoption label: the target was initially correct and changed to an incorrect peer answer under stance-only exposure. This label does not require self-reflection stability and therefore measures harmful peer adoption rather than strict conformity. Among 5,200 pairs, 501 are positive (9.63\%). The model achieves ROC-AUC 0.661 (question-cluster bootstrap 95\% CI [0.629, 0.692]) and PR-AUC 0.158 ([0.136, 0.187]), against a prevalence baseline of 0.096. Thus, the deployable features provide some risk-ranking signal, but their discrimination is limited and does not reliably determine whether a particular adoption will help or harm.

The diagnostic model ranks harmful conformity among 1,466 initially correct, self-reflection-stable targets, of whom 295 exhibit C$\to$W transitions (AUC = 0.784). Appendix~\ref{app:rq3_feature_ablation} shows that peer-pressure features alone match the full diagnostic model in AUC, while the largest wrong-answer coalition and peer support count rank above confidence variables. This diagnostic signal depends on ground-truth eligibility and is not deployable as an online correctness predictor.

\subsection{Diagnostic Intervention as Mechanism Check}
\label{sec:rq3_diagnostic}

We rank the 1,466 eligible samples by out-of-fold LR risk scores and rerun top-20\%, random-20\%, and bottom-20\% groups ($n=294$ each) with an independence-promoting intervention prompt.

Detailed diagnostic intervention outcomes are reported in Appendix~\ref{app:rq3_feature_ablation}, Table~\ref{tab:diagnostic-intervention} and Figure~\ref{fig:intervention}. In the high-risk group, the intervention reduces harmful conformity by 13.61 pp (95\% CI [7.5, 19.7]) and increases accuracy by 14.63 pp (95\% CI [8.2, 21.0]). Random and low-risk groups show no significant benefit. Eligibility and targeting use ground-truth correctness and self-reflection outcomes. The result therefore characterizes an oracle-eligible diagnostic setting, with deployment evaluated separately below.

\begin{table*}[t]
\centering
\caption{Deployment intervention changes (percentage points, pp) with paired question-cluster bootstrap 95\% CIs in brackets. H = harmful peer adoption (correct $\to$ wrong), B = beneficial peer adoption (wrong $\to$ correct), and W$\to$W = peer-adoption change among initially wrong targets that remain wrong. The H, B, and accuracy intervals use 2,000 paired resamples. The W$\to$W intervals use 5,000 resamples (seed 42).}
\label{tab:deployment_intervention}
\renewcommand{\arraystretch}{1.04}
\setlength{\tabcolsep}{3pt}
\small
\resizebox{\textwidth}{!}{%
\begin{tabular}{@{}llcccc@{}}
\toprule
\textbf{Policy} & \textbf{Scope} & \textbf{$\Delta$H} & \textbf{$\Delta$B} & \textbf{$\Delta$W$\to$W} & \textbf{$\Delta$Acc.} \\
\midrule
\multirow{2}{*}{Targeted} & top 20\% & $-$1.44 [$-$2.79, $-$0.10] & $-$2.12 [$-$4.33, 0.00] & $-$5.48 [$-$8.23, $-$2.80] & $-$0.77 [$-$3.46, $+$2.02] \\
 & full & $-$0.29 [$-$0.58, 0.00] & $-$0.42 [$-$0.87, $+$0.02] & $-$1.10 [$-$1.63, $-$0.56] & $-$0.15 [$-$0.73, $+$0.40] \\
\midrule
\multirow{2}{*}{Random} & rand 20\% & $-$0.67 [$-$2.31, $+$0.87] & $-$1.54 [$-$3.56, $+$0.58] & $-$1.44 [$-$3.44, $+$0.57] & $-$1.06 [$-$3.85, $+$1.83] \\
 & full & $-$0.13 [$-$0.46, $+$0.17] & $-$0.31 [$-$0.69, $+$0.08] & $-$0.29 [$-$0.67, $+$0.12] & $-$0.21 [$-$0.77, $+$0.33] \\
\bottomrule
\end{tabular}}
\end{table*}

\subsection{Deployment Setting}
\label{sec:rq3_deployment}

For deployment, we use the answer-agnostic stance-only peer-adoption score introduced in Section~\ref{sec:rq3_prediction} to rank all 5,200 agent--question pairs. We compare a policy targeting the top 20\% by out-of-fold risk with a random baseline, evaluating paired changes within the selected subsets and after applying each policy across the full sample.

Table~\ref{tab:deployment_intervention} reports the changes in three mutually exclusive transition groups and in accuracy. Under the targeted policy, harmful, beneficial, and wrong-to-wrong peer adoption all decrease. The targeted W$\to$W intervals exclude zero in both scopes, and the targeted harmful-adoption interval excludes zero in the top-20\% subset. Every accuracy interval includes zero, and no random-policy interval excludes zero. The adoption-risk ranking therefore identifies who is likely to change, but it does not distinguish beneficial peer adoption from harmful peer adoption. The deployment policy has no detectable accuracy benefit in this experiment. Practical deployment would require a correctness signal, a verifier, or another validated way to separate correction opportunities from harmful influence.
\par\medskip

\section{Discussion}
\label{sec:discussion}

This work shows why peer-associated changes must be separated by source and direction. Self-reflection accounts for a substantial share of observed changes, while strict-conformity events include harmful, beneficial, and wrong-to-wrong transitions. Their direction depends on the target's initial correctness and the answer supported by peers. Across the evaluated model--benchmark combinations, beneficial transitions are at least as frequent as harmful ones, although estimates outside the primary setting remain uncertain.

\paragraph{Why peer adoption does not always improve accuracy.}
Peer adoption can correct an answer only when a correct alternative is available. Among initially wrong, self-reflection-stable targets, 49.2\% have no correct peer and only 11.3\% face a correct-peer majority. Support-standardized beneficial adoption exceeds harmful adoption under both exposure conditions, while the oracle gradient shows that accuracy varies sharply with correct-peer availability. Raw conformity rates therefore conceal the opportunity structure that determines direction. The paired majority-vote transitions show that gains and losses can cancel at the decision level even when individual accuracy rises.

\paragraph{Responses to peer context.}
In the controlled 500-case experiment, constructed wrong-answer adoption increases with richer peer materials for GPT-4o, but the invalid-minus-assertion contrast reverses for DeepSeek-V4-Flash and is near zero for Qwen3-Plus. The more consistent increase from generic text to question-specific wrong rationales across all three targets suggests that plausible but false explanations can be influential. These complete materials differ in content, length, and presentation, so the contrast does not isolate explanation quality alone.

\paragraph{Implications for MAD design.}
MAD is most defensible when answer quality can be checked or unsupported answers can be independently verified. The diagnostic intervention improves accuracy in a selected high-risk group defined with ground-truth correctness and no-peer stability. The answer-agnostic deployment model instead ranks likely peer adopters, and its intervention suppresses both harmful and beneficial adoption without a detectable accuracy gain. Full reasoning also uses 57\% more tokens than self-reflection while yielding the same 77.35\% majority-vote accuracy. These results favor debate policies that can be validated against group-level decision quality and cost.

\section{Conclusion}
\label{sec:conclusion}

We introduced a matched decomposition of reconsideration instability, strict conformity, and reasoning-only peer adoption. Matched self-reflection, stance-only, and full-reasoning conditions supply a no-peer baseline for measuring peer-aligned adoption. Peer-associated changes can harm, correct, or preserve an incorrect answer, with direction governed by correction opportunity, peer support, and the target model's response to peer materials. Across evaluated settings, beneficial transitions are at least as frequent as harmful ones, though non-primary estimates remain uncertain. Although individual accuracy improves, full reasoning provides no additional majority-vote gain over self-reflection in the primary setting despite higher token cost. Suppressing peer adoption without a correctness signal also reduces both harmful and beneficial changes. Together, these findings support selective use of MAD when answer quality can be independently checked. Future work should test verification-guided policies across additional models, heterogeneous agent groups, and multi-round settings.

This study also has several limitations. The primary experiments use homogeneous GPT-4o agents with a single round of peer exposure, so heterogeneous groups and multi-round dynamics remain untested. The controlled RQ2 experiment uses susceptibility-selected anchors, assigned priors, and constructed peer materials, so its adoption rates are not population estimates for ordinary debate. All conclusions are further restricted to multiple-choice tasks with verifiable ground truth. Appendix~\ref{app:limitations} discusses these limitations in detail.

\section*{Ethics Statement}
All the experiments in this paper were conducted using LLM agents with no human subjects involved. 
To the best of our knowledge, there is no negative societal impact in this research. 
We used LLMs to polish the paper. 
We are responsible for all the materials presented in this work.



\section*{Acknowledgments}
This work was partially supported by NSFC 62472289, 62532007, 62502034, and
Guangdong Province Key Laboratory of Popular High Performance Computers
2017B030314073.

\bibliographystyle{iclr2027_conference}
\bibliography{anthology,custom}

\clearpage
\appendix
\section{Limitations}
\label{app:limitations}
Our study has several limitations. First, primary experiments use homogeneous GPT-4o agents. Selected DeepSeek-V4-Flash and Qwen3-Plus replications and cross-target RQ2 with fixed GPT-4o materials do not test native heterogeneous interaction, limiting generalization across capabilities and architectures. Second, we examine one peer-exposure round, leaving multi-round compounding or self-correction unexplored. Third, strict conformity requires stability in one self-reflection call and may miss adoption when no-peer responses are stochastic. Fourth, controlled RQ2 uses artificial invalid/non-diagnostic and wrong-reasoning materials, susceptibility-linked cases, one peer, and assigned priors differing from observed Round~0 for nearly half the anchors. Its rates are not population estimates for ordinary debate or natural peer reasoning. Finally, verifiable multiple-choice tasks may not reflect open-ended or subjective conformity dynamics.

\section{Robustness and Generalization}
\label{app:robustness}

These analyses extend RQ1 across benchmarks, model families, and accuracy--cost comparisons.

\subsection{Cross-Dataset Directional Analysis}
\label{app:gpqa_direction}

Table~\ref{tab:gpqa_direction} reports event-level transition counts for GPQA-Diamond using the same three-source decomposition as in the main text.

\begin{table}[h]
\centering
\caption{Directional transition counts for GPT-4o on GPQA-Diamond. Percentages are within each event cohort.}
\label{tab:gpqa_direction}
\begin{tabular}{lrrrr}
\toprule
\textbf{Event source} & $n$ & \textbf{C$\to$W (\%)} & \textbf{W$\to$C (\%)} & \textbf{W$\to$W (\%)} \\
\midrule
Strict conformity & 93 & 27.96 & 43.01 & 29.03 \\
Self-reflection changes & 206 & 28.16 & 31.07 & 40.78 \\
\bottomrule
\end{tabular}

\end{table}

\paragraph{Round-0 variability and re-elicitation sensitivity.}
We measured Round-0 variability with three identical information-isolated calls for each of the 198 GPQA-Diamond questions. The mean question-level pairwise disagreement rate was 22.4\% (95\% CI [17.8, 27.1]), and 62/198 questions (31.3\%) exhibited at least one disagreement.

The first Round-0 call for each question and repetition was used as a common anchor for two no-peer re-elicitation conditions. The answer-again condition used the neutral instruction ``Please answer this question again based on the above information,'' whereas the reconsideration condition used ``Please reconsider this question based on the above information.'' Both conditions provided the same prior answer, did not provide the prior reasoning, and differed only in the final instruction.

Across the 594 common anchors, answer-again changed 8.8\% of answers (95\% CI [5.9, 11.8]), while explicit reconsideration changed 16.7\% (95\% CI [13.3, 20.2]). The paired difference was 7.91 percentage points (question-clustered 95\% CI [4.38, 11.62]). The 39.24\% GPQA rate in Table~\ref{tab:flip_rates} uses 525 pairs from 105 Round-0 disagreement questions. The re-elicitation analysis covers all 198 questions, including unanimous questions, and therefore describes a different population and prompt comparison.

\subsection{Cross-Model Validation}
\label{app:cross_model}

To test whether the main findings are specific to GPT-4o, we replicate the experiment using DeepSeek-V4-Flash and Qwen3-Plus on both datasets. All models answer the same set of questions at Round~0 (4,900 for MMLU-Pro, 198 for GPQA-Diamond), but the number of disagreement questions entering Round~1 varies by model because models differ in their tendency to produce unanimous answers. For example, DeepSeek-V4-Flash produces disagreement on only 141 of 4,900 MMLU-Pro questions (yielding $n=705$ agent-question pairs), compared to 1,040 for GPT-4o ($n=5{,}200$) and 337 for Qwen3-Plus ($n=1{,}685$). Table~\ref{tab:cross_model} reports the results.

\begin{table*}[t]
\centering
\caption{Directional transition counts for strict conformity. The W$\to$C share is calculated among C$\to$W and W$\to$C events, with question-clustered 95\% bootstrap CIs. DeepSeek-V4 denotes DeepSeek-V4-Flash. MMLU-Pro and GPQA-Diamond are abbreviated as MMLU and GPQA.}
\label{tab:cross_model}
\setlength{\tabcolsep}{4pt}
\begin{tabular}{llrrrrl}
\toprule
\textbf{Dataset} & \textbf{Model} & $n$ & \textbf{C$\to$W} & \textbf{W$\to$C} & \textbf{W$\to$W} & \shortstack{\textbf{W$\to$C share (\%)}\\\textbf{[95\% CI]}} \\
\midrule
\multirow{3}{*}{MMLU} & GPT-4o & 960 & 295 & 349 & 316 & 54.2 [50.5, 57.9] \\
 & DeepSeek-V4 & 136 & 39 & 40 & 57 & 50.6 [37.7, 63.4] \\
 & Qwen3-Plus & 169 & 37 & 39 & 93 & 51.3 [38.0, 64.5] \\
\midrule
\multirow{3}{*}{GPQA} & GPT-4o & 93 & 26 & 40 & 27 & 60.6 [47.8, 73.9] \\
 & DeepSeek-V4 & 38 & 11 & 16 & 11 & 59.3 [39.3, 80.0] \\
 & Qwen3-Plus & 41 & 6 & 12 & 23 & 66.7 [47.0, 85.7] \\
\bottomrule
\end{tabular}

\end{table*}

In Table~\ref{tab:cross_model}, $n$ is the number of strict-conformity events, not the total Round~1 pair count. The corresponding Round~1 totals are 5,200/705/1,685 for GPT-4o/DeepSeek-V4-Flash/Qwen3-Plus on MMLU-Pro and 525/300/435 on GPQA-Diamond.

We also attempted replication with Llama-3-8B-Instruct but found that the disagreement rate was too low for meaningful analysis: only 67 of 4,900 MMLU-Pro questions and 3 of 198 GPQA-Diamond questions produced inter-agent disagreement at Round~0. The very low disagreement rate indicates limited response diversity under this temperature-0 API setting, leaving too few disagreement cases to analyze conformity behavior. We therefore exclude Llama-3-8B-Instruct from further analysis.

Across the six model-dataset combinations, W$\to$C counts are at least as large as C$\to$W counts. The corresponding W$\to$C share ranges from 50.6\% to 66.7\%. However, question-clustered intervals are wide for all but the primary GPT-4o MMLU-Pro estimate, so the direction pattern is descriptive outside the primary sample. The models also differ substantially in reconsideration stability: DeepSeek-V4-Flash has an 11.91\% self-reflection flip rate on MMLU-Pro, while Qwen3-Plus has a 74.07\% rate. These cross-model changes support reporting the decomposition components separately and do not establish a universal directional effect.

Two model-specific patterns deserve attention. First, the monotonic ordering of flip rates (SR $<$ stance $<$ reasoning) observed for GPT-4o does not hold universally. On MMLU-Pro, Qwen3-Plus's self-reflection flip rate (74.07\%) exceeds its stance-only flip rate (47.83\%). One possible explanation is that Qwen3-Plus is highly unstable under prompted reconsideration, and exposure to peer stances constrains the model's response variability, reducing overall answer volatility even as it introduces conformity pressure. This does not affect our strict conformity measure, which conditions on SR-stable samples, but it means that the monotonic flip-rate ordering reported for the primary GPT-4o setting in Section~\ref{sec:rq1} should not be read as a universal pattern.

Second, the relative frequency of self-reflection changes and peer-adoption events varies across models. In the primary GPT-4o MMLU-Pro sample, self-reflection changes are more frequent than strict conformity (1,923 vs. 960 events). The event counts reverse for DeepSeek-V4-Flash (84 vs. 136), whose self-reflection flip rate is low. This reinforces the need to distinguish event prevalence from transition direction and to condition conclusions on the model's stability profile.

\subsection{Cross-Dataset Validation}
\label{app:cross_dataset}

To test generalization beyond MMLU-Pro and GPQA-Diamond, we replicate the main experiment on BBEH and AGIEval using GPT-4o. BBEH is cited through its benchmark paper in the main text. For task-level provenance, its reused Geometric Shapes task derives from GeomVerse \cite{kazemi2023geomverse}, and its NYCC task derives from the New Yorker Caption Contest benchmark \cite{hessel2022androids}. Table~\ref{tab:cross_dataset} reports the results.

\begin{table*}[t]
\centering
\caption{Directional transition counts for GPT-4o strict conformity across four datasets. The W$\to$C share is calculated among C$\to$W and W$\to$C events, with question-clustered 95\% bootstrap CIs.}
\label{tab:cross_dataset}
\setlength{\tabcolsep}{5pt}
\begin{tabular}{lrrrrl}
\toprule
\textbf{Dataset} & $n$ & \textbf{C$\to$W} & \textbf{W$\to$C} & \textbf{W$\to$W} & \shortstack{\textbf{W$\to$C share (\%)}\\\textbf{[95\% CI]}} \\
\midrule
MMLU-Pro & 960 & 295 & 349 & 316 & 54.2 [50.5, 57.9] \\
GPQA-Dia. & 93 & 26 & 40 & 27 & 60.6 [47.8, 73.9] \\
BBEH & 188 & 42 & 53 & 93 & 55.8 [46.0, 65.2] \\
AGIEval & 165 & 45 & 47 & 73 & 51.1 [41.6, 60.9] \\
\bottomrule
\end{tabular}

\end{table*}

Across these four GPT-4o datasets, W$\to$C counts are at least as large as C$\to$W counts, including BBEH and AGIEval. The event counts and question-clustered uncertainty intervals are reported in Table~\ref{tab:cross_dataset}. Only the primary MMLU-Pro W$\to$C share interval excludes 50\%. The results extend the directional accounting across benchmarks, while further model and debate settings are needed to establish generality.

\subsection{Accuracy--Cost Trade-off}
\label{app:accuracy_cost}

For Table~\ref{tab:accuracy_cost}, single-instance accuracy is the mean across the five Round~0 responses (18,406/24,500), and single-instance token cost is the total Round~0 five-agent cost divided by five. Aggregation-only uses the Round~0 majority vote. Multi-instance token totals include all Round~0 calls and the applicable Round~1 calls on the 1,040 retained disagreement questions.

\subsection{Computational Budget}

The main GPT-4o experiments for RQ1--RQ3 cost approximately \$500. Additional robustness replications, ablation studies, and sensitivity analyses cost approximately \$700.

\section{Sensitivity Analyses}
\label{app:sensitivity}

We vary peer order, prompt wording, group size, and temperature on GPQA-Diamond to examine the design sensitivity of the rates in Section~\ref{sec:rq1}.

\paragraph{Answer order.}
Shuffling the order in which peer answers are presented changes the strict-conformity rate by less than 2 percentage points (29.15\% vs.\ 27.22\%), indicating minimal positional bias.

\paragraph{Prompt paraphrase.}
Rephrasing the debate prompt while preserving its semantic content yields a strict-conformity rate of 25.55\% (vs.\ 29.15\% in the main setting), with a corresponding decrease in stance flip rate from 46.48\% to 41.44\%. The qualitative pattern is preserved.

\paragraph{Agent count.}
Standard MAD operates under a fully connected communication topology: each agent sees all peers' responses, so context length grows linearly with group size. This creates a practical upper bound, as excessively long contexts degrade LLM performance \cite{liu-etal-2024-lost}. Existing MAD studies typically use 3--12 agents \cite{du2023improving,liang-etal-2024-encouraging,smit2024should,choi-etal-2025-empirical}. We test 3 to 11 agents (Table~\ref{tab:sensitivity}): strict-conformity rates range from 20.83\% to 32.60\% without a strictly monotonic trend, and the qualitative decomposition pattern persists across group sizes.

\paragraph{Agent count and accuracy.}
Table~\ref{tab:agent_count_accuracy} reports accuracy on the 60 disagreement questions common to all agent-count conditions, enabling a controlled comparison. Across most settings, Round~1 accuracy exceeds Round~0, especially under stance-only and full-reasoning exposure, but the gains are not monotonic across agent counts.

\begin{table}[h]
\centering
\caption{Accuracy (\%) by agent count on the 60 common disagreement questions (GPQA-Diamond, GPT-4o). R0 = Round~0, SR = self-reflection, Stance = stance-only, and Reas.\ = full reasoning.}
\label{tab:agent_count_accuracy}
\begin{tabular}{ccccc}
\toprule
\textbf{Agents} & \textbf{R0} & \textbf{SR} & \textbf{Stance} & \textbf{Reas.} \\
\midrule
3 & 29.4 & 32.2 & 31.7 & 35.6 \\
5 (main) & 32.3 & 33.7 & 40.7 & 39.3 \\
7 & 31.7 & 31.9 & 34.3 & 38.6 \\
9 & 32.6 & 34.3 & 35.0 & 38.7 \\
11 & 34.5 & 32.9 & 49.8 & 45.9 \\
\bottomrule
\end{tabular}

\end{table}

\paragraph{Temperature.}
Higher temperature increases reconsideration instability (self-reflection flip rate: 39\% $\to$ 47\%) but does not proportionally increase strict conformity (26--29\%). This pattern suggests that, in the evaluated settings, higher temperature increases self-reflection flip rates more clearly than strict-conformity rates.

\begin{table}[h]
\centering
\caption{Sensitivity analyses on GPQA-Diamond (GPT-4o, \%). The strict-conformity rate is calculated among self-reflection-stable samples.}
\label{tab:sensitivity}
\begin{tabular}{llcc}
\toprule
\textbf{Factor} & \textbf{Setting} & \textbf{Strict conf.} & \textbf{Self-reflection flip} \\
\midrule
\multirow{2}{*}{Order} & Original & 29.15\% & 39.24\% \\
 & Shuffled & 27.22\% & 39.81\% \\
\midrule
\multirow{2}{*}{Prompt} & Original & 29.15\% & 39.24\% \\
 & Paraphrased & 25.55\% & 34.64\% \\
\midrule
\multirow{5}{*}{Agents} & 3 & 20.83\% & 42.17\% \\
 & 5 (main) & 29.15\% & 39.24\% \\
 & 7 & 25.87\% & 38.97\% \\
 & 9 & 28.59\% & 37.49\% \\
 & 11 & 32.60\% & 38.47\% \\
\midrule
\multirow{3}{*}{Temp.} & 0 (main) & 29.15\% & 39.24\% \\
 & 0.5 & 25.91\% & 45.77\% \\
 & 1.0 & 27.01\% & 47.27\% \\
\bottomrule
\end{tabular}

\end{table}

\section{Correction Opportunity and Support Standardization}
\label{app:support_standardization}

We compare harmful peer adoption (H) among the 1,466 initially correct, reconsideration-stable pairs with beneficial peer adoption (B) among initially wrong, stable pairs that have at least one correct peer. Of 1,811 initially wrong stable pairs, 891 have no correct peer (49.2\% [46.9, 51.5]) and only 205 have a correct-peer majority (11.3\% [9.9, 12.9]). The remaining 920 form the B opportunity population. Within each exposure and target-support stratum, $H$ denotes harmful peer adoption by initially correct targets, whereas $B$ denotes beneficial peer adoption by initially wrong targets with at least one correct peer. Rates are standardized over support counts 0/1/2/3 using pooled H+B weights (11.6/19.2/25.8/43.5\%). For support stratum $s$, the weight is $w_s=(n_{H,s}+n_{B,s})/(1{,}466+920)$, where the stratum counts appear in Table~\ref{tab:support_strata}. Each question-cluster bootstrap replicate re-estimates the weights.

The standardized results are reported in the main text in Table~\ref{tab:support_standardized}. Table~\ref{tab:support_strata} provides the underlying unstandardized rates for each support stratum. This comparison conditions on correction opportunity and is distinct from the raw composition of observed peer-adoption events.

\begin{table}[H]
\centering
\caption{Unstandardized support-stratum rates underlying Table~\ref{tab:support_standardized}. Support denotes target-answer support among the four peers.}
\label{tab:support_strata}
\begin{tabular}{lrrrrr}
\toprule
\textbf{Exposure} & \textbf{Support} & \textbf{$n_H$} & \textbf{H (\%)} & \textbf{$n_B$} & \textbf{B (\%)} \\
\midrule
Stance-only & 0 & 98 & 68.4 & 178 & 60.1 \\
Stance-only & 1 & 238 & 41.6 & 219 & 47.9 \\
Stance-only & 2 & 362 & 22.4 & 253 & 34.0 \\
Stance-only & 3 & 768 & 6.3 & 270 & 18.9 \\
Full reasoning & 0 & 98 & 95.9 & 178 & 75.8 \\
Full reasoning & 1 & 238 & 61.3 & 219 & 73.1 \\
Full reasoning & 2 & 362 & 24.9 & 253 & 43.5 \\
Full reasoning & 3 & 768 & 6.5 & 270 & 14.8 \\
\bottomrule
\end{tabular}

\end{table}

\section{Oracle Correct-Signal Support Gradient}
\label{app:correct_signal_gradient}

Before observing outcomes, we identified 1,680 eligible initially wrong, reconsideration-stable GPT-4o pairs across 689 disagreement questions. With pre-specified SHA-256 seed 20260714, we selected one target per question and then 250 questions. Four separate stance-only calls share the same Round~0 anchor and fixed answers W1 (initial wrong), C (correct), W2 (primary wrong), and W3 (alternate wrong). Only peer composition changes. Peer positions are cyclically balanced (63/63/62/62), a Latin square assigns conditions to process/key slots, and calls are hash-interleaved. All 1,000 outputs are valid.

The first step preserves the 3:1 split and the W2 majority, replacing only W3 with C. Later contrasts jointly change correct-answer prevalence and support. The main-text Table~\ref{tab:correct_signal_gradient} reports the monotonic accuracy gradient. Because construction requires ground-truth labels, this is oracle-controlled mechanism evidence rather than a deployable policy.

\FloatBarrier
\section{Per-Category Breakdown}
\label{app:per_category}

\noindent Table~\ref{tab:exp1_per_category_breakdown} reports strict-conformity rates and self-reflection flip rates across MMLU-Pro subject categories. Its grouped header separates strict conformity from self-reflection changes for GPT-4o, DeepSeek-V4-Flash, and Qwen3-Plus.

\Needspace{0.40\textheight}
\begin{center}
\captionof{table}{Per-category breakdown of strict-conformity rates and self-reflection flip rates across MMLU-Pro subject categories. DeepSeek-V4 denotes DeepSeek-V4-Flash. GPT-4o and Qwen3-Plus are shown in full.}
\label{tab:exp1_per_category_breakdown}
\setlength{\tabcolsep}{2.5pt}
\renewcommand{\arraystretch}{1.08}
\begin{tabular}{lcccccc}
\toprule
 & \multicolumn{3}{c}{\textbf{Strict-conformity rate}} & \multicolumn{3}{c}{\textbf{Self-reflection flip rate}} \\
\cmidrule(r){2-4}\cmidrule(l){5-7}
\textbf{Category} & \textbf{GPT-4o} & \shortstack{\textbf{DeepSeek-}\\\textbf{V4}} & \shortstack{\textbf{Qwen3-}\\\textbf{Plus}} & \textbf{GPT-4o} & \shortstack{\textbf{DeepSeek-}\\\textbf{V4}} & \shortstack{\textbf{Qwen3-}\\\textbf{Plus}} \\
\midrule
biology & 33.3 & 15.4 & 30.0 & 33.5 & 13.3 & 84.6 \\
business & 26.4 & 21.1 & 34.7 & 36.5 & 5.0 & 77.7 \\
chemistry & 29.3 & 23.3 & 38.3 & 42.0 & 8.8 & 73.3 \\
computer science & 24.8 & 21.7 & 36.4 & 37.1 & 23.3 & 76.8 \\
economics & 25.6 & 30.8 & 48.0 & 32.1 & 48.0 & 64.3 \\
engineering & 31.0 & 20.0 & 65.5 & 38.2 & 7.1 & 67.8 \\
health & 29.5 & 22.7 & 38.5 & 32.1 & 2.2 & 76.4 \\
history & 27.5 & 24.0 & 26.7 & 24.3 & 0.0 ($n=25$) & 72.7 \\
law & 25.5 & 20.8 & 12.5 & 35.9 & 4.0 & 69.2 \\
math & 28.7 & 21.7 & 46.7 & 46.6 & 7.7 & 80.0 \\
other & 32.2 & 26.1 & 21.9 & 27.3 & 16.4 & 73.3 \\
philosophy & 30.5 & 18.4 & 50.0 & 37.0 & 24.0 & 81.5 \\
physics & 32.6 & 22.0 & 44.8 & 43.0 & 8.9 & 70.3 \\
psychology & 34.6 & 18.8 & 7.7 & 30.0 & 20.0 & 78.3 \\
\bottomrule
\end{tabular}
\end{center}
\section{Peer-Material Contrasts in an Outcome-Selected Subset}
\label{app:rq2_145}

To examine adoption after resistance to simpler peer exposure, we select 145 target-agent/question indices that retained an assigned correct prior under both self-reflection and stance-only exposure. This outcome-based rule fixes both baselines at zero wrong-answer selection in the conditional analysis below. The rates therefore describe responses conditional on retaining that prior and are reported separately from the primary 500-anchor experiment in Section~\ref{sec:rq2}. The assigned prior matches the target's observed Round~0 answer in 69 cases. An independent four-condition evaluation on the same indices examines sensitivity to prompts and peer materials.

\paragraph{Conditional information gradient.}

\begin{figure}[H]
\centering
\includegraphics[width=0.65\linewidth]{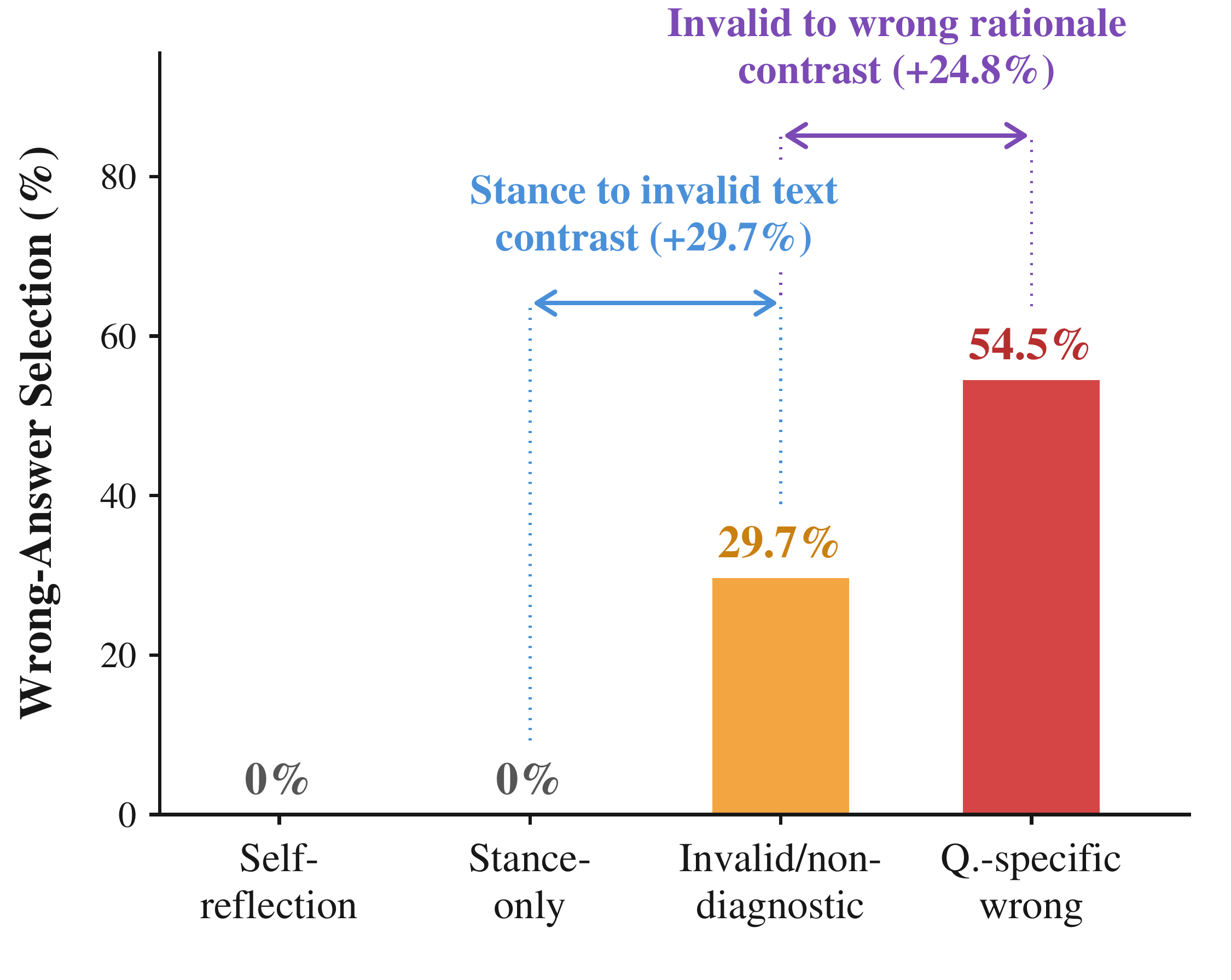}
\caption{Constructed wrong-answer selection across conditions in the outcome-selected $n=145$ analysis. The zero self-reflection and stance-only baselines follow from selecting cases that retained the assigned correct prior in both conditions. The annotations describe two incremental, bundled presentation contrasts.}
\label{fig:info_gradient}
\end{figure}

In this selected subset, Table~\ref{tab:control_highrisk} and Figure~\ref{fig:info_gradient} show a $+29.66$ pp increase from stance-only exposure to invalid/non-diagnostic reasoning and a further $+24.83$ pp increase with question-specific wrong reasoning. The conditions also differ in length, assertion, and argumentative formatting, so these estimates describe complete peer presentations.

\begin{table}[H]
\centering
\caption{Information-gradient analysis on the outcome-selected $n=145$ subset. Inv./non-diag.\ denotes generic non-solving text, and Q.-spec.\ wrong denotes a plausible question-specific incorrect rationale. Wrong-answer selection is the proportion of responses equal to the constructed wrong answer.}
\label{tab:control_highrisk}
\begin{tabular}{lrrr}
\toprule
\textbf{Condition} &
\shortstack{\textbf{Wrong-ans.}\\\textbf{sel.}} &
\textbf{95\% CI} &
\textbf{Acc.} \\
\midrule
Self-reflection & 0.00 & --- & 100.00 \\
Stance-only & 0.00 & --- & 100.00 \\
Inv./non-diag. & 29.66 & [22.8, 37.5] & 65.52 \\
Q.-spec. wrong & 54.48 & [46.4, 62.4] & 35.86 \\
\bottomrule
\end{tabular}

\end{table}

Because the peer materials are constructed, their wording and generator could shape adoption. Table~\ref{tab:exp2_groups} therefore varies the generic-text phrasing and, in one variant, uses Qwen3-Max instead of GPT-5.4 to generate the question-specific wrong rationale. The question-specific rationale elicits higher wrong-answer selection than generic non-solving text in each variant, while selection under generic text ranges from 20--39\%. The variants use separately selected subsets, so their rates do not isolate a generator effect or replace the pre-specified $n=500$ four-condition analysis.

\begin{table}[H]
\centering
\caption{Sensitivity analyses in selected subsets (\%). Phrasings~1--3 vary the generic non-solving text. Qwen3-Max replaces the question-specific wrong-reasoning generator.}
\label{tab:exp2_groups}
\begin{tabular}{llcc}
\toprule
\textbf{Variant} & \textbf{$n$} & \textbf{Inv./non-diag.} & \textbf{Q.-spec. wrong} \\
\midrule
Phrasing~1 & 145 & 29.66 & 54.48 \\
Phrasing~2 & 198 & 39.39 & 61.11 \\
Phrasing~3 & 128 & 20.31 & 52.34 \\
Qwen3-Max & 135 & 32.59 & 45.19 \\
\bottomrule
\end{tabular}

\end{table}

\paragraph{Four-condition evaluation with independent calls.}
We evaluated the same 145 selected indices using independent model calls and a four-condition design. All 145 cases were valid across 53 question clusters. Constructed wrong-answer selection was 45.5/48.3/50.3/72.4\% under the stance-only, assertion-only, invalid/non-diagnostic, and question-specific-wrong conditions, respectively. Invalid-minus-assertion was $+2.07$ pp [question-clustered 95\% CI $-4.03$, $+8.76$], whereas question-specific-wrong-minus-invalid was $+22.07$ pp [$+9.40$, $+35.66$]. The prompts and materials differ from the conditional analysis above and the primary $n=500$ experiment, so the rate levels are reported separately.

\section{Alternative 500-Anchor Peer-Material Protocol}
\label{app:rq2_original_full}

To examine protocol sensitivity in Section~\ref{sec:rq2}, we evaluate another 500-anchor information-gradient protocol with different model calls and peer materials from the primary four-condition experiment. Table~\ref{tab:control_full} reports its wrong-answer selection rates.

\begin{table}[h]
\centering
\caption{Alternative 500-anchor peer-material protocol (\%). The assigned correct prior matches the observed Round~0 answer in 253/500 cases. Wrong-ans.\ sel.\ denotes selection of the constructed wrong answer. $\Delta$Acc.\ is relative to the assigned-prior accuracy of 100\%. In self-reflection, the constructed wrong answer is not shown, so its selection is coincidental rather than peer adoption.}
\label{tab:control_full}
\begin{tabular}{lrrr}
\toprule
\textbf{Condition} & \textbf{Wrong-ans.\ sel.} & \textbf{Accuracy} & \textbf{$\Delta$Acc.} \\
\midrule
Self-reflection & 8.20 & 84.40 & $-$15.60 \\
Stance-only & 63.60 & 33.00 & $-$67.00 \\
Inv./non-diag. & 65.60 & 30.00 & $-$70.00 \\
Q.-spec. wrong & 77.80 & 17.40 & $-$82.60 \\
\bottomrule
\end{tabular}

\end{table}

Under this protocol, invalid/non-diagnostic text increases adoption by 2.0 pp over stance-only, while the question-specific wrong rationale increases it by 12.2 pp over invalid/non-diagnostic text. This protocol compares generic text with stance-only exposure, complementing the assertion-controlled contrast in the main experiment. The contrasts characterize sensitivity to the complete peer presentation.

\FloatBarrier
\section{Controlled Cross-Target RQ2 Extensions}
\label{app:cross_target_extensions}

DeepSeek-V4-Flash and Qwen3-Plus receive the same 500 GPT-4o Round~0 anchors, questions, options, constructed wrong answers, and GPT-generated materials as the primary experiment. Main-text Table~\ref{tab:rq2_cross_target_full} reports adoption rates for all three target models, and final accuracy is reported in Table~\ref{tab:rq2_cross_target_accuracy}. Thus, these are response-model substitutions from a fixed GPT-generated state, not native end-to-end or interactive heterogeneous MAD replications.

\begin{table}[H]
\centering
\caption{Final accuracy (\%) in the controlled cross-target four-condition experiment ($n=500$ per condition and target).}
\label{tab:rq2_cross_target_accuracy}
\begin{tabular}{lrrr}
\toprule
\textbf{Condition} & \textbf{GPT-4o} & \textbf{DeepSeek-V4} & \textbf{Qwen3-Plus} \\
\midrule
Stance-only & 32.8 & 77.6 & 73.6 \\
Assertion-only & 34.6 & 73.2 & 59.8 \\
Invalid/non-diagnostic & 20.8 & 85.8 & 60.0 \\
Question-specific wrong & 10.8 & 59.6 & 40.8 \\
\bottomrule
\end{tabular}
\end{table}

The assigned correct prior matches the target's actual Round~0 answer in 253/500 cases. Among the 247 mismatches, the target's actual wrong answer equals the constructed distractor in 193 cases. The remaining 54 have another wrong answer.

The assertion-to-invalid pattern is not uniform: adoption increases by 13.4 pp for GPT-4o, decreases by 12.0 pp for DeepSeek-V4-Flash, and changes by $-1.4$ pp for Qwen3-Plus. By contrast, question-specific-wrong minus invalid/non-diagnostic adoption is positive for all three targets ($+8.8/+25.4/+18.0$ pp). Paired contrasts and question-clustered confidence intervals are reported in main-text Table~\ref{tab:rq2_target_contrasts}. These are controlled response-model checks, not native end-to-end trajectories for DeepSeek-V4-Flash or Qwen3-Plus.

The invalid-minus-assertion sign reversal establishes target-dependent response to the bundled presentation manipulation. Wrong-minus-invalid is positive for all three targets, but prompt length and unmeasured presentation differences preclude a pure content-effect interpretation. Table~\ref{tab:rq2_target_interactions} quantifies the corresponding between-target interactions.

\begin{table}[H]
\centering
\caption{Assertion-to-invalid target interactions on the same 500 anchors. Whole-question bootstrap replicates retain paired observations for both targets.}
\label{tab:rq2_target_interactions}
\begin{tabular}{lc}
\toprule
\textbf{Target interaction} & \textbf{Difference-in-differences (pp)} \\
\midrule
GPT-4o $-$ DeepSeek-V4-Flash & $+$25.4 [$+$17.6, $+$33.8] \\
GPT-4o $-$ Qwen3-Plus & $+$14.8 [$+$7.4, $+$22.2] \\
Qwen3-Plus $-$ DeepSeek-V4-Flash & $+$10.6 [$+$3.4, $+$18.2] \\
\bottomrule
\end{tabular}

\end{table}

\section{Target Resistance under Identical Anonymous Pressure}
\label{app:target_resistance}

To complement the one-peer material comparisons in Section~\ref{sec:rq2}, we compare target resistance under identical majority pressure. We select agent-question pairs for which GPT-4o and Qwen3-Plus each answered correctly in their own Round~0. Both target models then receive the same stance-only context: three anonymous peers support one constructed wrong answer and one peer supports the correct answer. Model identities and capability labels are hidden.

Across 14,055 paired cases and 3,049 question clusters, Qwen3-Plus is 2.14 pp less likely than GPT-4o to abandon its correct answer [95\% CI $-$2.72, $-$1.54]. Within each target, abandoned-correct cases have lower standardized confidence than retained cases: event-minus-nonevent differences are $-$1.33 [$-$1.53, $-$1.14] for GPT-4o and $-$1.83 [$-$2.05, $-$1.61] for Qwen3-Plus. Table~\ref{tab:target_resistance} reports the paired target-resistance results. These within-model results do not justify cross-model confidence calibration. The experiment identifies target-dependent susceptibility under a constructed identical context, not capability effects or a full heterogeneous MAD pipeline.

\begin{table}[H]
\centering
\caption{Target resistance on paired common-valid cases under identical anonymous 3:1 wrong-majority pressure.}
\label{tab:target_resistance}
\begin{tabular}{lrrrr}
\toprule
\textbf{Target} & \textbf{$n$} & \textbf{Retain (\%)} & \textbf{Abandon (\%)} & \textbf{Adopt wrong maj. (\%)} \\
\midrule
GPT-4o & 14,055 & 94.95 & 5.05 & 4.05 \\
Qwen3-Plus & 14,055 & 97.09 & 2.91 & 1.38 \\
\bottomrule
\end{tabular}

\end{table}

\section{Diagnostic and Deployment Feature-Set Analysis}
\label{app:rq3_feature_ablation}

This appendix provides the intervention outcomes, confidence summaries, and feature-set analyses supporting Section~\ref{sec:rq3}.

\subsection{Diagnostic Intervention Results}

Table~\ref{tab:diagnostic-intervention} reports the complete diagnostic intervention results. The intervention targets the top-20\%, random-20\%, and bottom-20\% risk groups identified from the oracle-eligible samples.

\begin{table}[H]
\centering
\caption{Diagnostic intervention results (\%, $n=294$ per group). Flip denotes any answer change. C$\to$W denotes harmful conformity, conditional on self-reflection stability.}
\label{tab:diagnostic-intervention}
\begin{tabular}{llrrr}
\toprule
\textbf{Group} & & \textbf{Flip} & \textbf{C$\to$W} & \textbf{Acc.} \\
\midrule
\multirow{2}{*}{High} & Orig. & 57.48 & 52.72 & 42.52 \\
 & Interv. & 42.86 & 39.12 & 57.14 \\
\midrule
\multirow{2}{*}{Random} & Orig. & 25.51 & 21.77 & 74.49 \\
 & Interv. & 21.43 & 20.07 & 78.57 \\
\midrule
\multirow{2}{*}{Low} & Orig. & 10.88 & 7.14 & 89.12 \\
 & Interv. & 8.16 & 6.80 & 91.84 \\
\bottomrule
\end{tabular}

\end{table}

\begin{figure}[H]
\centering
\includegraphics[width=0.65\linewidth]{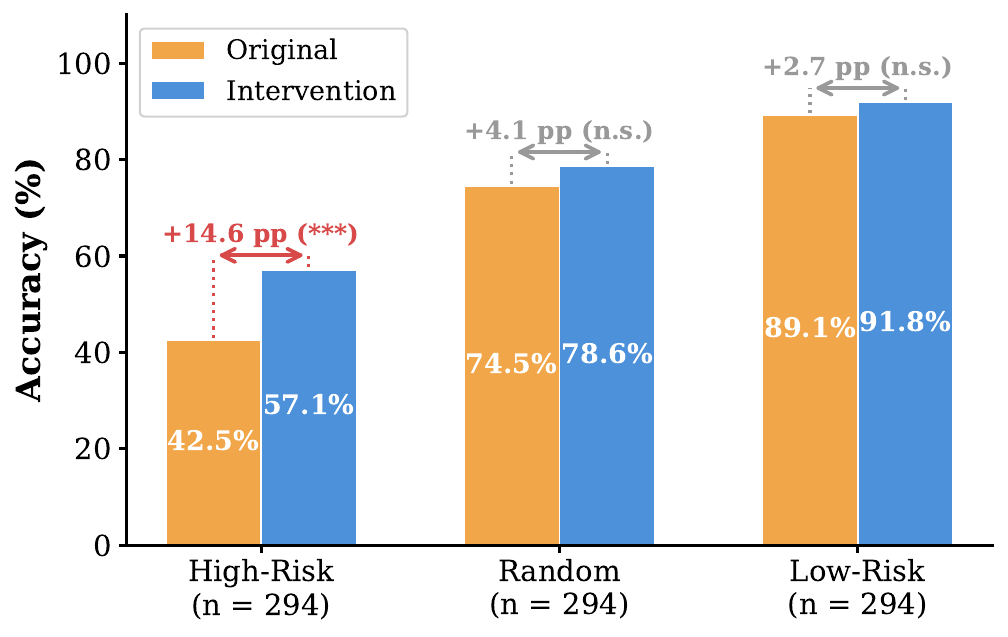}
\caption{Accuracy before and after intervention by risk group in the oracle diagnostic setting ($n=294$ per group). The plot shows point estimates. The question-clustered confidence interval for the high-risk-group change and the inferential interpretation of all three groups are reported in Section~\ref{sec:rq3}.}
\label{fig:intervention}
\end{figure}

\subsection{Confidence Summaries}

Round~0 confidence differs only modestly by correctness and behavioral outcome. The explicit- and implicit-confidence gaps are small relative to their within-group variation, which limits confidence as a standalone predictor of later answer changes.

\Needspace{0.25\textheight}
\par\medskip
\noindent
\begin{minipage}[t]{0.49\textwidth}
\vspace{0pt}
\centering
\captionof{table}{Round~0 confidence by correctness (MMLU-Pro, $24{,}500$ agent-question pairs).}
\label{tab:conf_correctness}
\renewcommand{\arraystretch}{0.98}
\setlength{\tabcolsep}{1pt}
\begin{tabular}{@{}lrcc@{}}
\toprule
& \textbf{$n$} & \textbf{Explicit} & \textbf{Implicit} \\
\midrule
Correct & 18406 & 99.45$\pm$2.16 & 91.69$\pm$15.94 \\
Incorrect & 6094 & 97.20$\pm$6.58 & 87.14$\pm$19.93 \\
\bottomrule
\end{tabular}
\end{minipage}
\hfill
\begin{minipage}[t]{0.49\textwidth}
\vspace{0pt}
\centering
\captionof{table}{Round~0 confidence by behavioral outcome (MMLU-Pro). The three reported groups total 4,570 pairs. The remaining 630 pairs comprise 487 reasoning-only peer adoptions and 143 ``Other'' cases and are not included in these summaries. ``Strict conformity'' denotes stance-only peer adoption after a stable self-reflection response.}
\label{tab:conf_behavior}
\renewcommand{\arraystretch}{0.98}
\setlength{\tabcolsep}{1pt}
\small
\begin{tabular}{@{}lrcc@{}}
\toprule
& \textbf{$n$} & \textbf{Explicit} & \textbf{Implicit} \\
\midrule
Reconsideration & 1923 & 96.36$\pm$8.93 & 84.49$\pm$22.21 \\
Strict conformity & 960 & 96.70$\pm$6.58 & 85.78$\pm$20.43 \\
Stable & 1687 & 97.37$\pm$5.31 & 88.62$\pm$18.10 \\
\bottomrule
\end{tabular}
\end{minipage}
\par\medskip

\FloatBarrier
\subsection{Feature-Set Analysis}

Table~\ref{tab:revised-feature-sets} reports the nested question-grouped cross-validation results. Diagnostic rows use oracle eligibility and the harmful-conformity label on 1,466 pairs, including 295 positive cases across 600 question clusters. The peer-adoption deployment row uses answer-agnostic Round~0 features to predict stance-only peer adoption on all 5,200 agent--question pairs, including 2,141 positive cases across 1,040 question clusters. The C$\to$W deployment row uses the same nine features and the same stance-only exposure, but labels initially correct targets that adopt an incorrect peer answer. It has 501 positive cases (9.63\%) across 1,040 question clusters and does not require self-reflection stability.

\Needspace{0.3\textheight}
\subsubsection{The Nine Deployable Round~0 Features}
\label{app:deployable_features}

\begin{enumerate}[label=(\arabic*),nosep,leftmargin=2em]
    \item Round~0 implicit confidence.
    \item Relative implicit confidence.
    \item Within-group implicit-confidence rank.
    \item Round~0 reasoning length.
    \item Number of peers sharing the target's answer.
    \item Target-is-alone indicator.
    \item Largest alternative peer-coalition size.
    \item Majority margin.
    \item Answer entropy.
\end{enumerate}
All nine features are computed from Round~0 target outputs and the distribution of peer answers, without ground-truth correctness or self-reflection outcomes.

\begin{table}[H]
\centering
\caption{Feature-set analysis under nested question-grouped cross-validation. Prev.\ denotes positive-class prevalence and ECE denotes expected calibration error computed with ten equal-width probability bins. Diagnostic and deployment tasks use different labels and evaluation populations, so their metrics are not directly comparable.}
\label{tab:revised-feature-sets}
\setlength{\tabcolsep}{1.8pt}
\begin{tabular}{p{0.14\textwidth}p{0.22\textwidth}rcccccc}
\toprule
Task & Feature set & $n$ & ROC-AUC [95\% CI] & PR-AUC & Prev. & Brier & ECE & F1 \\
\midrule
Diagnostic & Full diagnostic (8) & 1,466 & .784 [.745, .818] & .500 & .201 & .128 & .019 & .511 \\
Diagnostic & Peer-pressure only (4) & 1,466 & .789 [.754, .822] & .497 & .201 & .128 & .030 & .510 \\
Diagnostic & Non-pressure only (4) & 1,466 & .479 [.436, .519] & .195 & .201 & .163 & .026 & .326 \\
Diagnostic & No explicit coalition counts (5) & 1,466 & .651 [.609, .693] & .309 & .201 & .155 & .051 & .431 \\
Deployment & Deployable Round~0 (9) & 5,200 & .804 [.784, .820] & .738 & .412 & .173 & .016 & .695 \\
Deployment (C$\to$W) & Deployable Round~0 (9) & 5,200 & .661 [.629, .692] & .158 & .096 & .085 & .003 & .240 \\
\bottomrule
\end{tabular}

\end{table}

\subsubsection{Evaluation and Diagnostic Comparisons}

For each task and feature set, we report ROC-AUC, PR-AUC, positive-class prevalence, Brier score, ECE, and F1 from cross-fitted predictions. All preprocessing, feature scaling, regularization, and threshold selection are performed within the training folds of the nested question-grouped cross-validation. ECE is computed using ten equal-width probability bins. The reported ROC-AUC confidence intervals use 1,000 question-cluster bootstrap replicates. The remaining metrics are reported as point estimates from the cross-fitted predictions. Diagnostic and deployment AUCs are not directly comparable because their labels and evaluation populations differ. For the C$\to$W deployment row, the PR-AUC is 0.158 (95\% question-cluster bootstrap CI [0.136, 0.187], 1,000 resamples). The table reports its point estimate for consistency with the other rows.

The peer-pressure-only diagnostic feature set has a slightly higher point-estimate ROC-AUC than the full diagnostic feature set (0.789 vs.\ 0.784), although their 95\% confidence intervals overlap. This result is consistent with limited incremental predictive value from confidence and other non-pressure features in this setting.

\begin{figure}[H]
\centering
\includegraphics[width=0.50\linewidth]{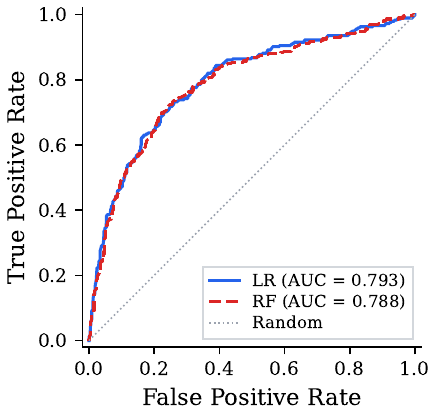}
\caption{Supplementary diagnostic model-comparison ROC curves under non-nested five-fold cross-validation ($n=1{,}466$). These descriptive AUCs are distinct from the nested question-grouped feature-set estimate reported in Table~\ref{tab:revised-feature-sets}.}
\label{fig:roc}
\end{figure}

Figure~\ref{fig:roc} reports a supplementary classifier comparison under non-nested five-fold cross-validation. Table~\ref{tab:feature_importance} summarizes feature importance, and Table~\ref{tab:ablation} reports supplementary logistic-regression feature expansions. The inferential estimates discussed in the main text use nested question-grouped cross-validation and are reported separately in Table~\ref{tab:revised-feature-sets}.

\begin{table}[H]
\centering
\caption{Diagnostic feature importance. LR = standardized logistic-regression coefficient, and RF = Gini importance.}
\label{tab:feature_importance}
\begin{tabular}{lcc}
\toprule
\textbf{Feature} & \textbf{LR} & \textbf{RF} \\
\midrule
Largest wrong-answer coalition & $+$.456 & .211 \\
Peer support count & $-$.409 & .296 \\
Answer entropy & $+$.192 & .127 \\
Reasoning length & $-$.150 & .084 \\
Target-is-alone indicator & $+$.092 & .142 \\
Implicit confidence & $-$.051 & .050 \\
\bottomrule
\end{tabular}

\end{table}

\begin{table}[H]
\centering
\caption{Supplementary logistic-regression feature-expansion analysis under five-fold cross-validation. These results are distinct from the nested question-grouped estimates in Table~\ref{tab:revised-feature-sets}.}
\label{tab:ablation}
\begin{tabular}{llcc}
\toprule
\textbf{Set} & \textbf{Added} & \textbf{AUC} & \textbf{F1} \\
\midrule
A: Base (8) & -- & .793 & .551 \\
B: +Ques.\ (10) & category, diff. & .797 & .544 \\
C: +Text (18) & hedging, TTR, \ldots & .792 & .543 \\
D: +Inter.\ (11) & conf$\times$pressure & .793 & .541 \\
E: All (23) & all above & .792 & .552 \\
\bottomrule
\end{tabular}

\end{table}

\FloatBarrier
\section{Implicit Confidence Details}
\label{app:implicit_confidence}

Sections~\ref{sec:methodology} and~\ref{sec:rq3} use implicit confidence derived from the token-level log probability of the generated final-answer token. This differs from the confidence value stated explicitly by the model in its response.

For each main generation call, we request token-level log-probability information using the following API settings:
\[
\begin{aligned}
\text{logprobs} &= \text{True},\\
\text{top\_logprobs} &= 1.
\end{aligned}
\]

After the model produces its response, we first parse the final answer option from the generated text, such as A, B, C, or D. We then search the corresponding log-probability data for the token matching this final answer. Since answer tokens may appear in different surface forms, such as ``A'', `` A'', or attached to nearby punctuation or text, the matching procedure normalizes tokens by stripping whitespace, converting letters to uppercase, and checking nearby token fragments when necessary.

If the token corresponding to the final answer is successfully identified, we read its log probability and convert it to a probability:
\[
p_{\mathrm{implicit}} = \exp(\ell_{\mathrm{answer}}),
\]
where \(\ell_{\mathrm{answer}}\) denotes the log probability of the generated answer token. The resulting value lies in the interval \([0,1]\). For reporting on the same 0--100 scale as the model's explicitly stated confidence, we optionally convert it to a percentage:
\[
c_{\mathrm{implicit}} = 100 \times p_{\mathrm{implicit}}.
\]

Thus, a higher implicit confidence indicates that the model assigned a larger generation probability to the final answer token, whereas a lower value indicates a weaker probabilistic preference for that answer.

Because the answer token is generated after the model's reasoning trace, this probability is conditional on the generated reasoning and preceding output. It should therefore not be interpreted as a pre-reasoning measure of uncertainty.

In the experimental records, the model's self-reported textual confidence is stored as explicit confidence, while the token-probability-based confidence is stored as implicit confidence. If the API does not return valid log-probability data, or if the final answer cannot be parsed reliably, the implicit confidence for that sample is recorded as missing. To ensure consistency between the recorded answer and its implicit confidence, when the system applies a secondary answer correction or normalization step, the implicit confidence is recomputed using the corrected final answer.

\section{Prompt Templates}
\label{app:prompt_templates}

\definecolor{PromptGreenBack}{RGB}{221,232,218}
\definecolor{PromptGreenFrame}{RGB}{90,120,96}

\subsection{Main Experiment}
\noindent The following prompt templates were used to elicit agents' initial information-isolated responses and their subsequent Round 1 revisions under different information-sharing conditions. Each template retains the same core task instruction and output format. The main variation lies in whether the agent receives only its own prior answer, its own prior answer together with other agents' answers, or additionally the reasoning traces of other agents. The prompts do not mention majority opinion, consensus, peer reliability, or the correctness of peer reasoning. Only the visible information channel changes across conditions. Agent identifiers used for logging and indexing are omitted from the templates below because they are not included in the messages sent to the model. The labels ``Agent~1,'' ``Agent~2,'' and so forth inside peer-information blocks are anonymous peer labels presented during Round~1.

\Needspace{0.25\textheight}
\captionof{table}{Prompt template for Round 0 information-isolated answering.}
\label{apptab:round0_prompt}
\begin{tcolorbox}[colback=PromptGreenBack, colframe=PromptGreenFrame, title = {\normalsize\bfseries Round 0 Information-Isolated Answer Prompt}, toptitle=1mm, bottomtitle=1mm, breakable]
You are an expert with professional knowledge in various fields. Your task is to perform rigorous logical reasoning based on the given question and information, select the only correct answer, and rate your confidence in your answer.

\vspace{\medskipamount}

Please solve the following question:

Question: \{question\}

\vspace{\medskipamount}

Options:

\{options\}

\vspace{\medskipamount}

Please answer in the following format:

[Reasoning]

[Position]

[Confidence]: (Please rate your certainty about this answer from 0--100, where 0 means completely uncertain and 100 means completely certain)

\end{tcolorbox}

\Needspace{0.25\textheight}
\captionof{table}{Prompt template for Round 1 self-reflection condition.}
\label{apptab:round1_self_reflection_prompt}
\begin{tcolorbox}[colback=PromptGreenBack, colframe=PromptGreenFrame, title = {\normalsize\bfseries Round 1 Self-Reflection Prompt}, toptitle=1mm, bottomtitle=1mm, breakable]
You are an expert with professional knowledge in various fields. Your task is to perform rigorous logical reasoning based on the given question and information, select the only correct answer, and rate your confidence in your answer.

\vspace{\medskipamount}

Please solve the following question:

Question: \{question\}

\vspace{\medskipamount}

Options:

\{options\}

\vspace{\medskipamount}

Your previous round answer: \{prev\_ans\}

\vspace{\medskipamount}

Please reconsider this question based on the above information.

\vspace{\medskipamount}

Please answer in the following format:

[Reasoning]

[Position]

[Confidence]: (Please rate your certainty about this answer from 0--100, where 0 means completely uncertain and 100 means completely certain)

\end{tcolorbox}

\Needspace{0.25\textheight}
\captionof{table}{Prompt template for Round 1 stance-only condition.}
\label{apptab:round1_stance_only_prompt}
\begin{tcolorbox}[colback=PromptGreenBack, colframe=PromptGreenFrame, title = {\normalsize\bfseries Round 1 Stance-Only Prompt}, toptitle=1mm, bottomtitle=1mm, breakable]
You are an expert with professional knowledge in various fields. Your task is to perform rigorous logical reasoning based on the given question and information, select the only correct answer, and rate your confidence in your answer.

\vspace{\medskipamount}

Please solve the following question:

Question: \{question\}

\vspace{\medskipamount}

Options:

\{options\}

\vspace{\medskipamount}

Your previous round answer: \{prev\_ans\}

\vspace{\medskipamount}

Other agents' answers:

Agent \{peer1\}'s answer: \{ans1\}

Agent \{peer2\}'s answer: \{ans2\}

Agent \{peer3\}'s answer: \{ans3\}

Agent \{peer4\}'s answer: \{ans4\}

\vspace{\medskipamount}

Please reconsider this question based on the above information.

\vspace{\medskipamount}

Please answer in the following format:

[Reasoning]

[Position]

[Confidence]: (Please rate your certainty about this answer from 0--100, where 0 means completely uncertain and 100 means completely certain)

\end{tcolorbox}

\Needspace{0.25\textheight}
\captionof{table}{Prompt template for Round 1 full reasoning condition.}
\label{apptab:round1_reasoning_prompt}
\begin{tcolorbox}[colback=PromptGreenBack, colframe=PromptGreenFrame, title = {\normalsize\bfseries Round 1 Reasoning Prompt}, toptitle=1mm, bottomtitle=1mm, breakable]
You are an expert with professional knowledge in various fields. Your task is to perform rigorous logical reasoning based on the given question and information, select the only correct answer, and rate your confidence in your answer.

\vspace{\medskipamount}

Please solve the following question:

Question: \{question\}

\vspace{\medskipamount}

Options:

\{options\}

\vspace{\medskipamount}

Your previous round answer: \{prev\_ans\}

\vspace{\medskipamount}

Other agents' answers:

Agent \{peer1\}'s answer: \{ans1\}

Agent \{peer2\}'s answer: \{ans2\}

Agent \{peer3\}'s answer: \{ans3\}

Agent \{peer4\}'s answer: \{ans4\}

\vspace{\medskipamount}

Other agents' reasoning:

Agent \{peer1\}'s reasoning: \{reason1\}

Agent \{peer2\}'s reasoning: \{reason2\}

Agent \{peer3\}'s reasoning: \{reason3\}

Agent \{peer4\}'s reasoning: \{reason4\}

\vspace{\medskipamount}

Please reconsider this question based on the above information.

\vspace{\medskipamount}

Please answer in the following format:

[Reasoning]

[Position]

[Confidence]: (Please rate your certainty about this answer from 0--100, where 0 means completely uncertain and 100 means completely certain)

\end{tcolorbox}

\subsection{Mitigated Intervention Experiment}

\noindent The mitigated intervention prompt was used in the risk-targeted intervention experiment. In this experiment, the prediction model was used as a risk-ranking model rather than as a hard classifier: samples with higher predicted probabilities of harmful conformity were prioritized for intervention. For the same stance-only setting used in the main experiment, the mitigated condition adds a short anti-conformity instruction before the agent revises its answer. The goal is not to prevent answer changes, but to require agents to change only when they can independently identify why their initial answer is wrong or incomplete and why an alternative answer is better.

\Needspace{0.25\textheight}
\captionof{table}{Prompt template for the mitigated intervention stance-only condition.}
\label{apptab:mitigated_intervention_prompt}
\begin{tcolorbox}[colback=PromptGreenBack, colframe=PromptGreenFrame, title = {\normalsize\bfseries Mitigated Intervention Prompt}, toptitle=1mm, bottomtitle=1mm, breakable]
You are an expert with professional knowledge in various fields. Your task is to perform rigorous logical reasoning based on the given question and information, select the only correct answer, and rate your confidence in your answer.

\vspace{\medskipamount}

Please solve the following question:

Question: \{question\}

\vspace{\medskipamount}

Options:

\{options\}

\vspace{\medskipamount}

Your previous round answer: \{prev\_ans\}

\vspace{\medskipamount}

Other agents' answers:

Agent \{peer1\}'s answer: \{ans1\}

Agent \{peer2\}'s answer: \{ans2\}

Agent \{peer3\}'s answer: \{ans3\}

Agent \{peer4\}'s answer: \{ans4\}

\vspace{\medskipamount}

Before revising your answer, first independently re-evaluate the question and your initial reasoning.

Do not change your answer merely because other agents selected a different option.

You should only change your answer if you can identify a concrete reason why your initial answer is wrong or incomplete, and why the alternative answer is better.

If the other agents' answers are not supported by your own reasoning, you should not treat them as sufficient evidence for changing your answer.

\vspace{\medskipamount}

Please reconsider this question based on the above information.

\vspace{\medskipamount}

Please answer in the following format:

[Reasoning]

[Position]

[Confidence]: (Please rate your certainty about this answer from 0--100, where 0 means completely uncertain and 100 means completely certain)
\end{tcolorbox}

\section{Invalid/Non-Diagnostic and Question-Specific Wrong-Reasoning Examples}
\label{app:reasoning_examples}

\noindent For the peer-material experiments in Section~\ref{sec:rq2}, we include three examples of invalid/non-diagnostic reasoning materials (generic non-solving text) and three examples of question-specific wrong-reasoning materials (plausible but incorrect rationales). These examples illustrate the distinction between generic text that does not diagnose the question and a substantive but erroneous inferential chain tied to the question.

\definecolor{ReasoningExampleGreen}{RGB}{238,246,235}
\definecolor{ReasoningRule}{RGB}{160,180,160}

\newtcolorbox{reasoningmaterialbox}[2][]{%
  breakable,
  colback=#2,
  colframe=ReasoningRule,
  boxrule=0.4pt,
  arc=0pt,
  left=2mm,
  right=2mm,
  top=1mm,
  bottom=1mm,
  before skip=2mm,
  after skip=2mm,
  #1
}

\Needspace{0.25\textheight}
\captionof{table}{Examples of invalid/non-diagnostic reasoning materials used in Experiment 2. These are generic non-solving texts.}
\label{tab:exp2_invalid_reasoning_examples}
\Needspace{6\baselineskip}
\begin{reasoningmaterialbox}[title={\normalsize\bfseries Invalid/non-diagnostic reasoning materials}]{ReasoningExampleGreen}
\textbf{Example 1: Generic suitability claim.}
\par \textbf{Reasoning:} After considering the question and reviewing all available options, this answer is a reasonable and appropriate choice. This option fits the overall context of the question, and there is no clear reason to exclude it. Since the selected answer best matches the expected response, it should be regarded as the most suitable answer. In other words, this answer is correct because it provides the kind of conclusion required by the question.

In addition, compared with the other options, this choice better aligns with the overall direction of the question. The reason for selecting this answer is that a good answer should be consistent with the main idea, and this option exactly matches that main idea. Therefore, based on the overall structure of the question and the suitability of this option, this answer should be selected.
\end{reasoningmaterialbox}

\Needspace{6\baselineskip}
\begin{reasoningmaterialbox}{white}
\textbf{Example 2: Relevance without evidence.}
\par \textbf{Reasoning:} When evaluating the question, the selected answer appears to be a generally reasonable choice. It follows the overall direction of the question and provides a response that seems consistent with what the question is asking for. Among the available options, this choice appears to fit the expected form of the answer more closely than the alternatives.

Moreover, a correct answer should usually align with the main purpose of the question and lead to a clear conclusion. The selected option satisfies this requirement because it does not seem to conflict with the general context of the problem. Therefore, based on its overall relevance, consistency, and suitability, this answer should be regarded as the most appropriate choice.
\end{reasoningmaterialbox}

\Needspace{6\baselineskip}
\begin{reasoningmaterialbox}{ReasoningExampleGreen}
\textbf{Example 3: Structural completeness claim.}
\par \textbf{Reasoning:} Overall, this answer demonstrates a reasonable structure and exhibits a degree of logical coherence. The option follows general evaluation criteria and aligns with the broader framework of the question. Compared to other options, it appears more complete in form, and can therefore be considered an acceptable choice.

Furthermore, a suitable answer should typically conform to the overall requirements of the question and present a clear conclusion. This option does not show any evident flaws in these respects and does not deviate from the general direction of the problem. Consequently, considering its logical consistency and structural completeness, this answer can be selected.
\end{reasoningmaterialbox}

\vspace{\baselineskip}

\Needspace{0.25\textheight}
\captionof{table}{Examples of question-specific wrong-reasoning materials used in Experiment 2. These texts contain plausible but incorrect rationales tied to the question.}
\label{tab:exp2_wrong_reasoning_examples}
\Needspace{6\baselineskip}
\begin{reasoningmaterialbox}[title={\normalsize\bfseries Question-specific wrong-reasoning materials}]{ReasoningExampleGreen}
\textbf{Example 1: GARCH features (financial econometrics).}
\par \textbf{Selected answer:} (ii) and (iii) only.
\par \textbf{Reasoning:} A standard GARCH(1,1) model is designed primarily to describe how volatility evolves over time, so the most natural features it captures are leverage-type asymmetries in practice and volatility clustering.

First, volatility clustering, item (iii), is a classic strength of GARCH models. The conditional variance depends on past squared shocks and past variance, which means periods of high volatility tend to be followed by high volatility, and calm periods tend to persist as well. This directly matches the observed clustering in financial returns.

Second, item (ii), leverage effects, can also be associated with GARCH-style volatility modeling because negative return shocks often lead to larger subsequent volatility than positive shocks of similar size. Even though the basic framework is variance-driven, it is commonly used to reflect this asymmetric market behavior through its dependence on recent return innovations.

By contrast, item (i), fat tails in the unconditional return distribution, is not something a standard GARCH(1,1) automatically guarantees unless one additionally assumes a non-normal error distribution such as Student-t. And item (iv) is not captured because standard GARCH models volatility conditional on returns, rather than modeling volatility as directly determining the mean return process.

So the best choice is (ii) and (iii) only.
\end{reasoningmaterialbox}

\Needspace{6\baselineskip}
\begin{reasoningmaterialbox}{white}
\textbf{Example 2: Sinusitis mechanism (clinical medicine).}
\par \textbf{Selected answer:} Immune system disorder.
\par \textbf{Reasoning:} The most likely underlying mechanism is an immune system disorder.

Several clues suggest that this is not just a routine upper respiratory problem. She has involvement of multiple adjacent structures: maxillary sinus tenderness, referred pain to the upper incisors, middle ear effusion behind the tympanic membrane, tonsillar enlargement, and impaired frontal sinus transillumination. That pattern points to poor clearance of mucosal secretions across the upper respiratory tract rather than a single localized trigger.

Although she is young and otherwise appears healthy, immune abnormalities can first become apparent in adolescence or early adulthood, especially milder humoral defects that predispose patients to recurrent or unusually extensive infections of the sinuses, ears, and throat. The absence of smoking, allergies, or other chronic risk factors makes an intrinsic host defense problem more plausible. In addition, her symptoms have persisted for a week and progressed despite what sounds like a mild initial illness, which suggests ineffective immune containment.

An immune system disorder would allow normal respiratory pathogens to persist and spread through contiguous mucosal surfaces, producing sinus obstruction, middle ear fluid, and ongoing inflammation. This best explains the breadth and progression of her presentation.
\end{reasoningmaterialbox}

\Needspace{6\baselineskip}
\begin{reasoningmaterialbox}{ReasoningExampleGreen}
\textbf{Example 3: Attorney-client privilege (legal reasoning).}
\par \textbf{Selected answer:} A woman visited an attorney to discuss her plans for starting a new business. She shared her innovative business model and strategies with the attorney. The attorney later used this information to start a similar business of his own.
\par \textbf{Reasoning:} Option B is the strongest choice because the communication took place in the classic attorney-client setting: a woman intentionally sought legal advice from an attorney and disclosed sensitive information about a proposed business venture. Attorney-client privilege generally protects confidential communications made for the purpose of obtaining legal counsel, and her discussion of the business model and strategies clearly fits that purpose.

What matters is the nature of the conversation at the time it occurred, not the attorney's later misconduct. The attorney's decision to use the information to launch a similar business does not change the privileged character of the original exchange. Privilege exists to encourage clients to speak openly with counsel, especially when discussing plans, risks, and legal structure for a new enterprise. If such conversations were not privileged, clients would hesitate to disclose essential details needed for meaningful advice.

By contrast, several other options involve public settings, third-party exposure, crime or fraud, or communications outside a true legal consultation. In B, the client met privately with an attorney, shared confidential business information, and did so specifically to obtain legal guidance. That is exactly the kind of communication privilege is designed to cover.
\end{reasoningmaterialbox}

\end{document}